\documentclass{ieeeaccess}

\usepackage{cite}
\usepackage{amsmath,amssymb,amsfonts}
\usepackage{graphicx}
\usepackage{textcomp}
\usepackage[utf8]{inputenc}
\usepackage[T1]{fontenc}
\usepackage{subcaption}
\usepackage{soul}
\usepackage{algorithm}
\usepackage{algpseudocode}
\usepackage{booktabs}
\usepackage{array}
\usepackage{tabularx}
\usepackage{caption}
\usepackage{float}
\usepackage{url} 
\usepackage{hyperref}
\usepackage{xcolor}
\newcommand{\pt}{(\ding{51})}
\newcommand{\grouphead}[1]{\addlinespace[2pt]\multicolumn{9}{@{}l}{\textit{#1}}\\[1pt]}
\hypersetup{
    colorlinks=true,
    urlcolor=blue
}
\usepackage{pifont}
\newcommand{\yes}{\ding{51}}
\newcommand{\no}{\ding{55}}
\newcolumntype{Y}{>{\centering\arraybackslash}X}
\newcolumntype{L}{>{\raggedright\arraybackslash}X}

\def\BibTeX{{\rm B\kern-.05em{\sc i\kern-.025em b}\kern-.08em
    T\kern-.1667em\lower.7ex\hbox{E}\kern-.125emX}}

\begin{document}
% ============================================================

\history{Date of publication xxxx 00, 0000, date of current version xxxx 00, 0000.}
\doi{10.1109/ACCESS.2017.DOI}

\title{Safe Multi-Robot Coordination via VLM-LLM Reasoning and Reachability Analysis}

\author{%
\uppercase{Mohamed Dwedar*}\authorrefmark{1,2},
\uppercase{Ahmad Hafez*}\authorrefmark{1},
\uppercase{Alexander Jesser}\authorrefmark{2},
\uppercase{Amr Alanwar}\authorrefmark{1}%
}

\address[1]{TUM School of Computation, Information, and Technology (CIT), Technical University of Munich (TUM), Heilbronn, Germany.}
\address[2]{Institute for Intelligent Cyber-Physical Systems (ICPS), Heilbronn University of Applied Science, Heilbronn, Germany.}

\tfootnote{This work is funded by the Deutsche Forschungsgemeinschaft (DFG, German Research Foundation) – Project number: 564740977.}

\markboth
{Dwedar, Hafez \headeretal: Safe Coordination of Heterogeneous Multi-Robot Systems}
{Dwedar,Hafez \headeretal: Safe Coordination of Heterogeneous Multi-Robot Systems}
\corresp{$^{*}$Contributed equally to this work. Corresponding authors: Mohamed Dwedar, Ahmad Hafez (e-mail: mohamed.dwedar@tum.de,a.hafez@tum.de).}

%\corresp{Corresponding author (Contributed Equally to this work: Mohamed Dwedar, Ahmad Hafez (e-mail: a.hafez@tum.de).}

% ============================================================
\begin{abstract}Safe coordination in heterogeneous machine-to-machine (M2M) robotic systems is difficult when robots differ in sensing capability, environmental awareness, and motion execution roles. This study presents a centralized safety aware M2M framework for cooperative goal-directed navigation in a heterogeneous mobile robot system composed of a vision-capable robot and a cameraless robotic vehicle. The objective is to guide the robots toward a detected goal region, such as a traversable target area or open door direction, while avoiding static and dynamic obstacles and preventing unsafe inter robot interactions. The central cooperation principle is shared perception: the vision-capable robot provides semantic environmental awareness through a centralized server, allowing the camera-less robot to act using this shared scene representation together with its own odometry, IMU, and state feedback. Both robots communicate with the server through an MQTT broker and continuously publish robot-state data, while the vision capable robot additionally transmits visual observations. A vision language model interprets the scene, and the extracted semantic information is converted into conservative geometric constraints, including obstacle regions, traversable areas, goal regions, safe corridors, and motion boundaries. A large language model supports high level task allocation reasoning by proposing robot specific navigation decisions, while deterministic controllers remain responsible for low-level execution. Before any command is issued, each candidate action is verified by a zonotope based reachability engine that checks obstacle avoidance, safe corridor containment, and inter-robot collision constraints. Only commands satisfying these reachability-based safety conditions are approved and transmitted to the corresponding robot. Online validation in clear-path and dynamic-obstacle scenarios demonstrates that the framework can approve safe motion, trigger conservative replanning or holding behavior, and preserve a strict separation between semantic reasoning and executable control. The proposed framework unifies shared semantic perception, broker-based M2M communication, cooperative task allocation, and formal reachability verification to support safer coordination of heterogeneous mobile robots with asymmetric sensing capabilities.\end{abstract}

\begin{keywords}
Machine-to-Machine Communication,
Heterogeneous Robotic Systems,
Multi-Robot Coordination,
Vision-Language Models,
Large Language Models,
Zonotope Reachability Analysis,
Robotic Safety,
Shared Perception,
MQTT,
Safe Autonomy
\end{keywords}

\titlepgskip=-15pt
\maketitle

% ============================================================
\section{Introduction}
\label{sec:introduction}
% ============================================================

\PARstart{T}{he} integration of mobile robots, embedded sensors,
edge computing nodes, and shared communication infrastructure has
established machine-to-machine~(M2M) interaction as a foundational
paradigm in cyber-physical and robotic systems~\cite{donta2022iotprotocols}.
In these architectures, platforms exchange state telemetry, perception
data, and task-level decisions through lightweight messaging protocols,
enabling cooperative operation without synchronous command
channels~\cite{mishra2020mqtt}. This paradigm is especially relevant
in heterogeneous robot teams, where platforms differ in sensing
capability, computational resources, mobility, and operational
role~\cite{mayya2021resilient,queralta2020sar}. However, communication
infrastructure alone does not ensure safe coordination. Shared robot
states and perception streams must be validated temporally, aligned
geometrically, and verified against motion-safety constraints before
they translate into executable commands~\cite{verma2021multirobot,quinton2023mrta}.
A robot that receives a position estimate or scene description from a
network peer must confirm that the data is fresh, geometrically
consistent with its own state, and physically feasible before treating
it as a basis for motion. The information pipeline from perception to
action must therefore include not only a communication layer but also
a formal verification stage that gates command issuance on measurable
safety conditions.

The coordination challenge intensifies under asymmetric sensing, where
one robot carries an onboard camera while another operates without
local visual perception. A camera-less robot cannot directly observe
obstacles, free space, goal candidates, or dynamic elements in its
environment. Without shared perception from a sensing partner, it
must navigate under conservative blind assumptions that substantially
limit its usefulness in goal-directed tasks. This asymmetry is not a
degenerate edge case: practical deployments often pair capable
perception platforms with cost-constrained or lightweight vehicles
that carry minimal sensor suites. For such configurations, shared
perception through a reliable M2M pipeline is structurally necessary
for informed coordination. However, relaying visual data from one
robot to another introduces synchronization requirements, coordinate
transformation challenges, and latency that must be accounted for in
the safety analysis. Critically, the camera-less robot must use the
shared scene information together with its own proprioceptive state
feedback, and every derived command must be verified against that
robot's own kinematic constraints and current pose, not those of the
perception source.

Vision-language models~(VLMs) and large language models~(LLMs) have
demonstrated substantial capability in semantic scene interpretation,
task decomposition, and high-level action selection for robotic
systems~\cite{wang2025llmrobotics,jeong2024llmrobot,fan2025vlmhrc,han2026robotvisionvlm}.
Embodied-AI systems have shown that visual and language reasoning can
support instruction-following navigation and multi-step robotic
planning~\cite{driess2023palme,zitkovich2023rt2,ahn2023saycan}.
Despite these advances, semantic reasoning models are not suitable
substitutes for formal motion-safety verification. Foundation models
may produce plausible but dynamically infeasible action proposals,
may fail to account for platform-specific kinematic constraints, or
may overlook collision geometry in scenes that differ from their
training distribution~\cite{huang2023innermonologue,huang2023voxposer,liang2023codeaspolicies}.
Additionally, VLM and LLM inference latency, output stochasticity,
and the absence of explicit dynamic models make these components
unsuitable for operation within high-frequency low-level control
loops. A responsible integration strategy must therefore treat VLM
and LLM outputs as structured advisory proposals subject to
downstream formal verification, maintaining a strict architectural
separation between semantic reasoning and motor command authority.

Reachability analysis provides a computationally tractable framework
for formal online safety verification in robotic systems subject to
bounded disturbances, parametric uncertainty, and admissible input
sets~\cite{althoff2021setpropagation,althoff2018cora,chen2018hjreachability}.
By propagating sets of states forward in time, reachability methods
bound the region of space a robot may occupy over a finite prediction
horizon, enabling safety predicates to be evaluated against obstacle
sets, corridor boundaries, and inter-robot separation
constraints~\cite{kochdumper2021sparse,liu2024refine}.
Zonotopes are particularly effective for this purpose. An affine map
applied to a zonotope produces another zonotope, and the Minkowski
sum of two zonotopes is itself a zonotope, both computable in closed
form. These properties enable exact propagation of reachable sets
through the linearized dynamics without numerical integration, at a
computational cost that scales polynomially with the number of
generators~\cite{kousik2020rtd,michaux2023rdf,michaux2024sparrows}.
Zonotope methods have been applied to data-driven safety verification,
online recursive state estimation, and barrier certificate
synthesis~\cite{alanwar2022data,alanwar2023data,oumer2025data}.
More recently, reachability has been proposed as a supervisory gate
to filter language-model action proposals for single-robot
systems~\cite{hafez2025safe}; this paper extends that principle to
a heterogeneous two-robot team with asymmetric sensing.

This paper presents a centralized safety-filtering pipeline that
integrates MQTT-based M2M communication, shared VLM-based perception,
LLM-supported symbolic task reasoning, semantic-to-geometric constraint
conversion, and zonotope reachability verification for cooperative
coordination of a heterogeneous robot team. The team consists of a
Unitree~Go2 quadruped, serving as the vision-capable platform, and
an SVEA robot car operating without an onboard
camera~\cite{jiang2022svea}. Both robots publish odometry, IMU, and
state messages to a centralized server through an MQTT broker; the
Go2 additionally streams live camera frames. The server interprets
the visual stream using Qwen2.5-VL via~Ollama, converts the validated
semantic output into conservative metric constraints, and then uses
Qwen3 via~Ollama to propose symbolic task-level actions. These
proposals are treated as candidate inputs to a Python zonotope
reachability module, which propagates robot-specific reachable tubes
and evaluates obstacle avoidance, safe-corridor containment, and
inter-robot separation predicates before any command is published.
The camera-less SVEA benefits from Go2-derived shared perception
while contributing its own synchronized proprioceptive state and
receiving independently verified robot-specific commands.
\subsection{Method at a Glance}
\label{subsec:glance}

\begin{figure}[h]
\centering
\includegraphics[width=0.5\textwidth,keepaspectratio]{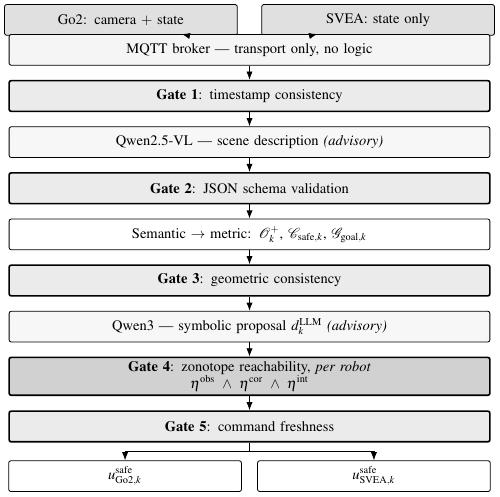}
\caption{Simplified view of the proposed method. Shaded blocks are gates;
light blocks are advisory and carry no command authority. Semantic evidence
produced by the Go2 propagates to SVEA, but command authority does not:
Gate~4 is evaluated independently for each robot.}
\label{fig:simplified}
\end{figure}

Figure~\ref{fig:simplified} gives a simplified view of the method before the
formal treatment in Section~\ref{sec:methodology}. The pipeline is a five-gate
cascade. Sensing from both robots enters over MQTT and must first pass a
timestamp-consistency gate. The surviving packet is interpreted by the VLM,
whose JSON output must pass a schema gate. The validated description is
projected onto the ground plane and converted into metric constraint sets,
which must pass a geometric-consistency gate. The LLM then proposes one
symbolic action. That proposal is not a command: it only selects which
candidate control is submitted to the fourth gate, the zonotope reachability
engine, evaluated \emph{once per robot} using that robot's own state and input
bounds. Only if all three reachability predicates hold does a command packet
reach the fifth gate, a freshness check, and only then is it published. Failure
at any gate yields a conservative \textit{HOLD} or \textit{STOP}. The single
structural idea the figure conveys is that semantic evidence flows across robots while command authority never does.

The main contributions of this paper are:
\begin{enumerate}
\item A centralized M2M coordination architecture for heterogeneous
robots with asymmetric sensing, in which a vision-capable Unitree~Go2
provides shared semantic--geometric perception to a camera-less SVEA
platform over an MQTT broker, while both platforms actively contribute
synchronized proprioceptive state feedback to every decision cycle.

\item A semantic-to-geometric conversion module that maps structured
VLM scene outputs into conservative metric obstacle sets, inflated
unsafe regions, safe corridors, and goal regions through camera
calibration, homography projection, robot-footprint modeling, and
Minkowski obstacle inflation parameterized by localization, perception,
communication, and conservative safety-margin uncertainty components.

\item A safety-filtered task reasoning pipeline in which Qwen3
proposes symbolic task-level actions from validated scene context but
cannot authorize motor commands; all proposals are treated as candidate
inputs subject to downstream zonotope reachability verification.

\item A robot-specific zonotope reachability gate that propagates
independent reachable tubes for Go2 and SVEA over a short finite
horizon using an affine unicycle model, and evaluates obstacle
avoidance, safe-corridor containment, and inter-robot separation
predicates independently for each robot before MQTT command
publication.

\item An online experimental validation under controlled indoor
laboratory conditions covering clear-path and dynamic-human-obstacle
scenarios, with demonstrated conservative rejection under
synchronization failure, VLM schema inconsistency, and reachability
predicate violation, together with a per-cycle representative latency
budget analysis.
\end{enumerate}
A video demonstration of the proposed approach is available online\footnote{\url{https://youtu.be/rnQ_K9NX4EU}}, and our results can be reproduced using our publicly available repository\footnote{\url{https://github.com/TUM-CPS-HN/MRSVLMRA}}.

\subsection{What Is Adopted, What Is Adapted, and What Is New}
\label{subsec:oldnew}
Because the pipeline integrates several established techniques, it is
important to state which parts are inherited and which are original.
 No novelty is
claimed for MQTT transport, for the zonotope algebra, for CORA-style set
propagation, for the unicycle abstraction, or for the two off-the-shelf
language models: all are used unmodified.

Two elements are adaptations. The reachability-as-supervisor principle is due
to Hafez~\textit{et~al.}~\cite{hafez2025safe} and applies to a single robot
verified against its own sensing; adapting it here required making the gate
\emph{robot-indexed}, so that one shared constraint set is evaluated against
$|\mathcal{R}|$ distinct reachable tubes rather than one. Minkowski obstacle
inflation is likewise standard; the adaptation is the decomposition of
$r_{\mathrm{safe}}$ in Eq.~\eqref{eq:inflation} into physically attributable
components including a perception term driven jointly by VLM confidence and
ground-plane reprojection error, a decomposition that
Section~\ref{subsec:stat_uncertainty} shows to be quantitatively necessary
rather than merely prudent.

Three elements are new. First, shared-perception constraint transfer: geometric
constraints derived from one robot's camera become the verification basis for a
robot that has no camera, with the receiving robot's own state parameterizing
its tube. Second, the inter-robot separation predicate $\eta^{\mathrm{int}}$
evaluated on paired tubes, which has no analogue in single-robot reachability
filters. Third, the coupling of the command validity horizon to the end-to-end
latency, $T_{\mathrm{valid}} \ge T_{\mathrm{total}} + \delta_{\mathrm{margin}}$,
which makes staleness a first-class safety condition rather than a networking
detail. We agree that several individual components are incremental; the claim
defended here is that the \emph{combination} one robot's perception becoming
another robot's formally verified constraint set is absent from both adjacent
literatures, as quantified in Section~\ref{subsec:positioning}.

The remainder of this paper is organized as follows.
Section~\ref{sec:related_work} reviews M2M communication, heterogeneous shared
perception, foundation-model robotics, and set-based safety verification, and
positions the pipeline against thirteen representative systems in
Table~\ref{tab:sota_comparison}. Section~\ref{sec:methodology} formalizes the
architecture: state and stream synchronization, semantic interpretation,
semantic-to-geometric conversion, LLM symbolic reasoning, the reduced dynamics
and their linearization, zonotope tube propagation, the three safety
predicates, and command publication under a freshness condition.
Section~\ref{sec:experimental_validation} describes the platform configuration,
implementation stack, online decision loop, and the test matrix in
Table~\ref{tab:test_matrix}. Section~\ref{sec:results} reports the quantitative
results, the statistical analysis of the uncertainty model, the comparison
against the state of the art, and a discussion of generalizability to
real-world deployment. Section~\ref{sec:latency} analyzes the per-cycle latency
budget and its relation to the command validity horizon.
Section~\ref{sec:ablation} presents the baseline and ablation protocol.
Section~\ref{sec:limitations} states the limitations bounding the reported
claims, and Section~\ref{sec:conclusion} concludes with quantified outcomes and
directions for future work.

% ============================================================
\section{Related Work}
\label{sec:related_work}
% ============================================================

\subsection{M2M Communication and Multi-Robot Coordination}

Lightweight M2M messaging protocols have been widely adopted to
support asynchronous, low-overhead telemetry exchange between
distributed robotic
agents~\cite{mishra2020mqtt,banno2021mqttloader,mileva2021mqtt5,seoane2021coapmqtt}.
MQTT in particular has been studied for throughput, latency, and
reliability under varying network configurations in IoT and robotic
deployments~\cite{donta2022iotprotocols,quincozes2024xaiot,shvaika2025tbmq}.
These studies confirm that local broker configurations can support the
low-latency telemetry exchange rates required in robotic coordination
tasks. However, existing M2M communication works predominantly address
the transport and reliability properties of the messaging layer itself;
they do not address how perception streams should be shared across
heterogeneous platforms, how shared scene representations should be
converted into safety constraints, or how command publication should
be gated on formal safety verification. In the proposed pipeline,
MQTT is deliberately confined to a transport role, carrying robot
state messages, camera streams, and approved command packets without
performing any planning, reasoning, or verification. This design
prevents the broker from becoming a safety-critical component and
ensures that communication-layer failures result in conservative
command rejection rather than unsafe command issuance.

\subsection{Heterogeneous Robots and Shared Perception}

Heterogeneous robot teams have been studied in search-and-rescue,
surveillance, and cooperative mapping scenarios where platforms with
different sensor suites and mobility profiles cooperate toward shared
objectives~\cite{queralta2020sar}. Resilient task allocation in
heterogeneous teams requires mechanisms that account for differing
observability and platform
capability~\cite{mayya2021resilient,verma2021multirobot,quinton2023mrta}.
Shared-perception approaches relay visual or semantic information
from a sensor-capable platform to one that lacks local sensing, but
most existing designs assume symmetric communication, do not address
the coordinate transformation and synchronization challenges involved,
or do not integrate formal safety verification for the receiving
platform. The SVEA testbed, used as the camera-less mobile platform in
this work, has been demonstrated in prior work as a capable
nonholonomic vehicle for experimental coordination
research~\cite{jiang2022svea}. In the proposed pipeline, SVEA is
explicitly not treated as a passive receiver of shared data: it
contributes synchronized odometry, IMU, and state feedback at every
decision cycle, and every command it receives is verified using its
own kinematic state, constraint set, and reachable set independently
from the Go2.

\subsection{VLM and LLM Reasoning for Robotics}

Foundation models have been applied to a wide range of robotic
perception and planning tasks. SayCan~\cite{ahn2023saycan} grounds
LLM plan proposals in robot affordances; PaLM-E~\cite{driess2023palme}
integrates language and visual tokens for embodied multi-step
reasoning; RT-2~\cite{zitkovich2023rt2} uses vision-language
co-training to improve robotic action generation. In multi-robot
settings, language models have been used for coalition formation and
task decomposition~\cite{kannan2024smartllm,wang2025llmrobotics}.
VLMs have been applied to human-robot collaboration scenarios where
visual context informs sequential action
selection~\cite{fan2025vlmhrc,han2026robotvisionvlm,jeong2024llmrobot}.
Despite these advances, foundation models cannot reliably enforce
kinodynamic constraints, quantify propagated uncertainty over a
prediction horizon, or guarantee collision-free motion in scenes
that differ from their training
distribution~\cite{huang2023innermonologue,huang2023voxposer,liang2023codeaspolicies}.
Hafez~\textit{et~al.}~\cite{hafez2025safe} propose a reachability
gate to filter single-robot LLM proposals; the proposed pipeline
extends this paradigm to a heterogeneous two-robot team with
structurally asymmetric sensing, requiring shared perception from
one robot to the other, robot-specific reachable-set evaluation, and
MQTT-based safety-filtered command publication.

\subsection{Reachability Analysis and Zonotope-Based Safety Verification}

Set-based reachability analysis computes the states reachable under
bounded inputs and disturbances and has been applied to safety
verification, control synthesis, and collision
avoidance~\cite{althoff2021setpropagation,althoff2018cora,chen2018hjreachability}.
Zonotopes support closed-form propagation under affine maps and
Minkowski sums, making them computationally tractable for online
predicate evaluation~\cite{kochdumper2021sparse,kochdumper2023safesafe}.
Real-time deployment~(RTD)~\cite{kousik2020rtd} and related
methods~\cite{michaux2023rdf,michaux2024sparrows,liu2024refine} apply
forward reachability to safe trajectory tracking and motion planning.
In data-driven settings, zonotopes have supported recursive state
estimation and safety verification under bounded
noise~\cite{alanwar2022data,alanwar2023data,oumer2025data}. These
works establish the formal set-propagation foundations used in this
paper, but they do not address heterogeneous multi-robot coordination
under asymmetric sensing, the integration of language-model-generated
symbolic proposals, or the connection between VLM semantic confidence
and conservative geometric safety margins.

\subsection{Research Gap and Positioning}
\label{subsec:positioning}
Existing works address M2M communication, heterogeneous robot
coordination, VLM/LLM reasoning, and zonotope reachability
verification in largely separate research contexts. Communication
works do not integrate semantic safety constraint construction.
Language-model works do not enforce kinodynamic reachability
constraints. Reachability works do not consider shared perception
asymmetry or language-model-generated symbolic proposals. In contrast,
this paper integrates shared VLM-based scene interpretation,
LLM-supported symbolic coordination, MQTT-based M2M communication,
and zonotope reachability verification into a single online
safety-filtered pipeline for heterogeneous robot coordination with
asymmetric sensing. The key distinction is that the Go2's visual
perception directly informs the geometric safety constraints used to
evaluate and approve SVEA's commands, while SVEA's own synchronized
state feedback parameterizes its independent robot-specific
reachable-set evaluation at every cycle.

\begin{table*}[!t]
\caption{Capability comparison against representative systems from the three
literatures the proposed pipeline draws on. \yes{}~provided; \pt{}~partial,
offline, or informal; \no{}~not provided. The final row tallies how many of the
twelve prior systems cover each axis.}
\label{tab:sota_comparison}
\centering
\footnotesize
\setlength{\tabcolsep}{4pt}
\renewcommand{\arraystretch}{1.1}
\begin{tabularx}{\textwidth}{@{}l l cccccc L@{}}
\toprule
& & \multicolumn{6}{c}{\textbf{Capability axis}} & \\
\cmidrule(lr){3-8}
\textbf{System} & \textbf{Category} &
\textbf{SEM} & \textbf{SYM} & \textbf{GATE} & \textbf{MR} &
\textbf{XFER} & \textbf{INT} &
\textbf{Safety mechanism reported} \\
\midrule
\grouphead{Language-model planning and vision--language--action models}
SayCan~\cite{ahn2023saycan} & LLM planning &
\yes & \yes & \no & \no & \no & \no &
Affordance scoring; no kinodynamic verification. \\
PaLM-E~\cite{driess2023palme} & VLA model &
\yes & \yes & \no & \no & \no & \no &
End-to-end; safety implicit in training data. \\
RT-2~\cite{zitkovich2023rt2} & VLA model &
\yes & \pt & \no & \no & \no & \no &
Co-trained action tokens; no runtime gate. \\
Code-as-Policies~\cite{liang2023codeaspolicies} & LLM code generation &
\yes & \yes & \no & \no & \no & \no &
Programmatic guards written by the model itself. \\
VoxPoser~\cite{huang2023voxposer} & VLM value maps &
\yes & \pt & \pt & \no & \no & \no &
Cost-map avoidance; not a set-based guarantee. \\
Inner Monologue~\cite{huang2023innermonologue} & LLM feedback &
\yes & \yes & \no & \no & \no & \no &
Textual replanning on failure detection. \\
LM-Nav~\cite{shah2023lmnav} & VLN navigation &
\yes & \pt & \no & \no & \no & \no &
Topological graph search; no reachability check. \\
\grouphead{Multi-robot language-model coordination}
SMART-LLM~\cite{kannan2024smartllm} & Multi-robot LLM &
\yes & \yes & \no & \yes & \no & \no &
Task decomposition and coalition formation only. \\
Chen \textit{et al.}~\cite{chen2024scalablemultirobot} & Multi-robot LLM &
\pt & \yes & \no & \yes & \no & \no &
Dialogue-based conflict resolution; no formal gate. \\
\grouphead{Set-based reachability planning}
RTD~\cite{kousik2020rtd} & Reachability planning &
\no & \no & \yes & \no & \no & \no &
Forward-reachable-set trajectory guarantee. \\
REFINE~\cite{liu2024refine} & Reachability planning &
\no & \no & \yes & \no & \no & \no &
Parameterized reachable sets for vehicles. \\
SPARROWS~\cite{michaux2024sparrows} & Reachability planning &
\no & \no & \yes & \no & \no & \no &
Spherical reachable sets for manipulators. \\
\grouphead{Language model with a formal gate}
Hafez \textit{et al.}~\cite{hafez2025safe} & LLM $+$ reachability &
\yes & \yes & \yes & \no & \no & \no &
Zonotope gate filtering single-robot LLM proposals. \\
\midrule
\textbf{This work} & \textbf{VLM$+$LLM$+$reachability} &
\yes & \yes & \yes & \yes & \yes & \yes &
Per-robot zonotope tubes; three predicates; latency-coupled command expiry. \\
\midrule
\multicolumn{2}{@{}l}{\textit{Coverage among the 12 prior systems}} &
7/12 & 6/12 & 6/12 & 2/12 & 0/12 & 0/12 & \\
\bottomrule
\end{tabularx}

\vspace{3pt}
\begin{minipage}{\textwidth}\footnotesize
\textbf{SEM}~semantic or language-conditioned scene input;\;
\textbf{SYM}~symbolic or task-level reasoning;\;
\textbf{GATE}~formal runtime safety gate on the issued command;\\
\textbf{MR}~heterogeneous multi-robot team;\;
\textbf{XFER}~constraint transfer to a platform lacking exteroceptive sensing;\;
\textbf{INT}~inter-robot separation predicate on reachable tubes.
\end{minipage}
\end{table*}

Table~\ref{tab:sota_comparison} makes this positioning quantitative rather
than qualitative. The axes are deliberately binary and verification-oriented
rather than performance-oriented, because the systems differ in task, platform
and metric: SayCan reports plan success on a mobile manipulator, RTD reports
tracking error on a Segway, and no published system reports a command-approval
rate for a camera-less robot verified against a partner's perception. A single
accuracy figure spanning such heterogeneous baselines would misrepresent all of
them.

At the capability level the comparison is unambiguous. Of the twelve prior
systems, $7/12$ provide semantic or language-conditioned scene input, $6/12$
provide symbolic or task-level reasoning, and $6/12$ provide a formal runtime
safety gate but only one~\cite{hafez2025safe} combines language-model
reasoning with a formal gate, and it does so for a single robot.
Two~\cite{kannan2024smartllm,chen2024scalablemultirobot} address multi-robot
teams, but neither verifies proposals against kinodynamic reachable sets.
\emph{No} prior system satisfies either of the two axes that define the present
problem: constraint transfer to a platform without exteroceptive sensing
($0/12$) and an inter-robot separation predicate on paired reachable tubes
($0/12$). The proposed pipeline satisfies $6/6$. The gap is therefore
structural rather than incremental: the reachability literature verifies robots
against their own sensors, and the language-model robotics literature shares
semantics but not verification.

% ============================================================
\section{Methodology}
\label{sec:methodology}
% ============================================================

\begin{figure*}[h]
\centering
\includegraphics[width=0.78\textwidth,keepaspectratio]{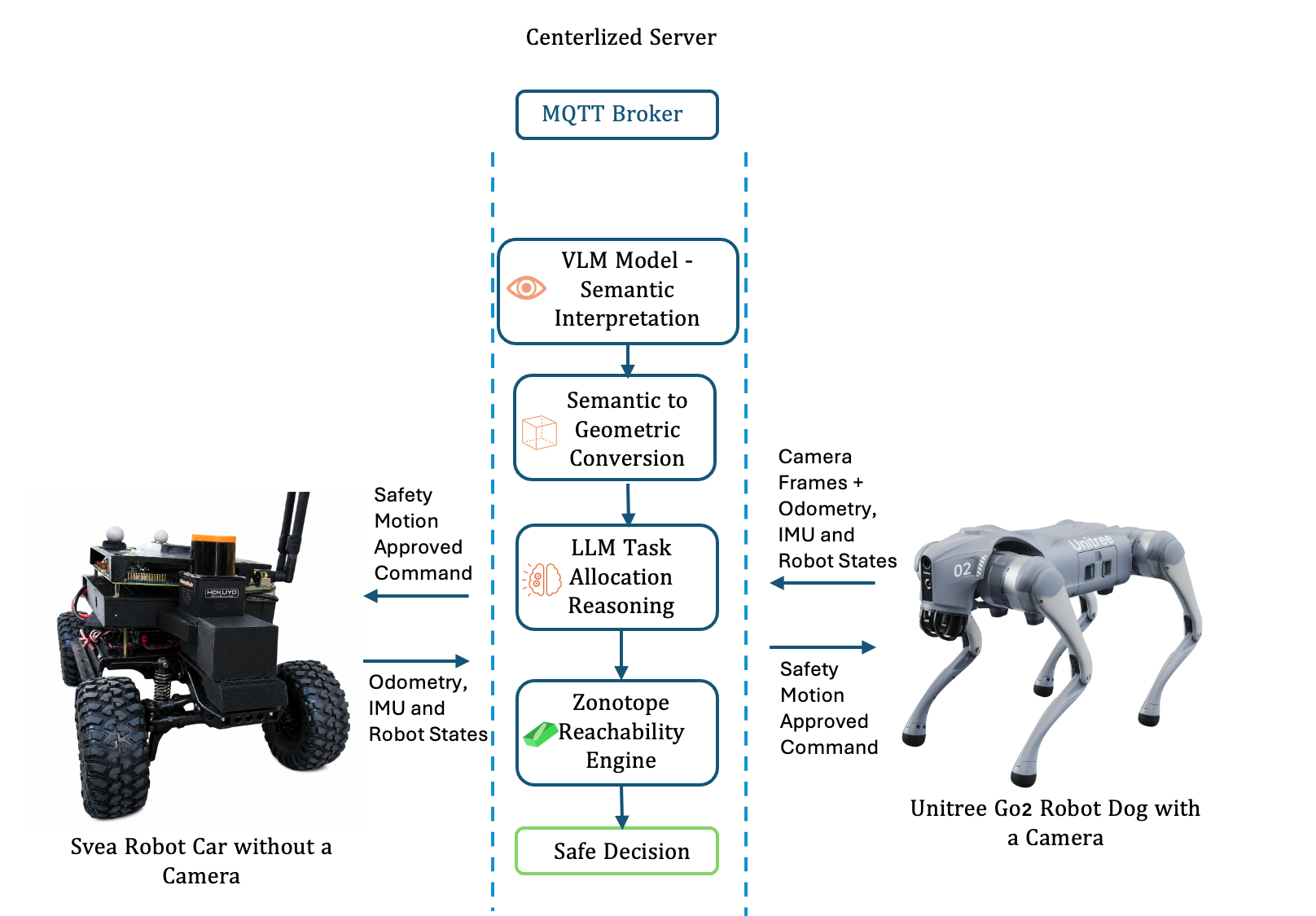}
\caption{Centralized M2M multi-robot safety decision pipeline.}
\label{fig:M2M}
\end{figure*}

Figure~\ref{fig:M2M} summarizes the centralized safety-aware
coordination pipeline. The system comprises a camera-less SVEA robot
car and a vision-capable Unitree~Go2 quadruped. Both platforms
transmit odometry, IMU, and state data to a centralized server
through an MQTT broker; the Go2 additionally publishes live camera
frames. The server synchronizes these heterogeneous inputs,
interprets the visual stream semantically, converts the semantic
output into conservative metric geometric constraints, evaluates a
symbolic task-level reasoning proposal, and verifies candidate
commands through zonotope reachability analysis before transmitting
robot-specific approved commands. MQTT functions exclusively as an
asynchronous M2M transport layer. It carries telemetry, camera
streams, and approved command packets but performs no planning,
reasoning, or verification~\cite{mishra2020mqtt,donta2022iotprotocols,shvaika2025tbmq}.
The communication layer is deliberately kept separate from the
decision-making layer so that a broker failure, network delay, or
packet loss results in a conservative \textit{HOLD} directive rather
than in an unsafe command.

\subsection{Novelty Delineation Within the Methodology}
The individual mechanisms below are drawn from established literature. The methodological
contribution is not any one mechanism but three specific couplings between
them, each of which appears in this section. First, the constraint sets
produced by Eq.~\eqref{eq:geometry} from the \emph{Go2} camera are used to
verify commands for the \emph{SVEA}, while Eq.~\eqref{eq:reach} is instantiated
with SVEA's own state, input set and disturbance envelope; this decoupling of
constraint source from verification subject is what makes a camera-less
platform usable without blind conservatism. Second, the predicate
$\eta^{\mathrm{int}}$ in Eq.~\eqref{eq:safety_predicates} couples two
independently propagated tubes, a condition that does not exist in single-robot
reachability filters. Third, the freshness condition in
Eq.~\eqref{eq:freshness} binds command validity to the measured pipeline
latency of Section~\ref{sec:latency}, so that staleness is enforced as a safety
predicate rather than handled as a transport-layer concern. Readers primarily
interested in what is new may read Sections~\ref{subsec:oldnew},
\eqref{eq:safety_predicates} and~\eqref{eq:freshness} together.

\subsection{Robot State and Stream Synchronization}

To enable robot-specific safety verification, the state of each
robot must be precisely defined. At time step~$k$, the state of
robot~$i$ is
\begin{equation}
\begin{aligned}
x_i(k) &=
\bigl[p_{x,i}(k),\; p_{y,i}(k),\; \theta_i(k),\; v_i(k)\bigr]^{\top}, \\
q_i(k) &=
\bigl[p_{x,i}(k),\; p_{y,i}(k),\; \theta_i(k)\bigr]^{\top},
\end{aligned}
\label{eq:state}
\end{equation}
where $p_{x,i}$ and $p_{y,i}$ are the planar Cartesian coordinates,
$\theta_i$ is the heading angle, and $v_i$ is the forward velocity.
The full state $x_i(k)$ is used to initialize the candidate control
bounds, while the reduced configuration $q_i(k) \in \mathbb{R}^3$ is
used for reachability analysis. This reduction is justified because
all three safety predicates obstacle avoidance, corridor
containment, and inter-robot separation depend on the predicted
planar occupancy of the robot, which is fully determined by
$(p_x, p_y, \theta)$. Heading is retained because it governs the
orientation of the robot footprint during set expansion. Velocity is
treated implicitly through the admissible input set and disturbance
bounds rather than as an explicit state to be checked against spatial
predicates. This model is not a full mechanical reproduction of the
Go2 or SVEA platforms; it is an intentionally reduced representation
for short-horizon planar safety envelope estimation.

Because Go2 and SVEA publish data at different rates and over
potentially variable network paths, the server must determine whether
the data streams are sufficiently aligned before using them jointly
for safety verification. The server accepts a synchronized packet
only if all required stream timestamps satisfy
\begin{equation}
\bigl|t_k^{a} - t_k^{b}\bigr| \leq \epsilon_t, \quad
a,b \in \{\mathrm{cam},\;\mathrm{odom},\;\mathrm{imu},\;\mathrm{state}\},
\label{eq:sync}
\end{equation}
where $\epsilon_t$ is the admissible synchronization tolerance.
Packets that violate this condition are discarded and the server
issues a conservative \textit{HOLD} or \textit{STOP} directive.
The synchronization check is safety-critical for the following
reason: if a stale camera frame is paired with a current SVEA state
packet, the obstacle positions computed from the image no longer
correspond to the physical locations of objects in the workspace
at the time the SVEA command would be executed. The resulting
safety assessment would be incorrect, potentially approving a command
that leads to a collision. By discarding mismatched packets before
any reasoning or verification, the pipeline ensures that every safety
check reflects a consistent snapshot of the physical system.

\subsection{Semantic Scene Interpretation}

The synchronized Go2 camera frame is processed locally by Qwen2.5-VL
running via~Ollama to generate a structured semantic description of
the workspace~\cite{fan2025vlmhrc,han2026robotvisionvlm}. The VLM
output is parameterized as
\begin{equation}
\begin{aligned}
S_k &= \{s_j(k)\}_{j=1}^{N_k}, \\
s_j(k) &= \bigl(c_j,\; b_j,\; \alpha_j,\; t_k\bigr),
\end{aligned}
\label{eq:semantic}
\end{equation}
where $c_j$ is the semantic class label of entity~$j$, $b_j$ is its
image-space bounding polygon, $\alpha_j$ is the model confidence
score, and $t_k$ is the associated timestamp. The confidence score
is an explicit field in the semantic representation rather than a
threshold applied at the VLM output stage. This design choice is
deliberate: a high-confidence detection can still exhibit significant
ground-plane reprojection error if the object geometry is complex or
the camera angle is shallow; conversely, a moderate-confidence
detection may project accurately when the geometry is simple. Because
$\alpha_j$ alone cannot determine the appropriate spatial safety
buffer, it must be combined with the geometric reprojection quality
at the conversion stage. Schema validation is applied to every VLM
JSON output before it propagates further; malformed, incomplete, or
internally inconsistent outputs are discarded and do not reach the
geometric conversion or reachability stages.

\subsection{Semantic-to-Geometric Conversion}

The validated semantic output $S_k$ contains image-space
descriptions of scene elements, but the zonotope reachability gate
requires metric geometric constraints in the robot workspace frame.
The mapping from semantic to geometric representation is
\begin{equation}
\label{eq:geometry}
\begin{aligned}
\mathcal{G}_{k}
&= \Phi\!\left(
S_{k}, T_{\mathrm{cam}}, K,
\mathcal{H}_{\mathrm{fp}}, x_{\mathrm{robot}}
\right) \\
&= \left\{
\mathcal{O}_{k}, \mathcal{O}_{k}^{+},
\mathcal{C}_{\mathrm{safe},k},
\mathcal{G}_{\mathrm{goal},k},
\pi_{k}
\right\}.
\end{aligned}
\end{equation}
where $T_{\mathrm{cam}}$ is the camera pose relative to the robot,
$K$ is the camera intrinsic matrix, $\mathcal{H}_{\mathrm{fp}}$ is
the floor-plane homography used to project image-space regions onto
the ground plane, $\mathcal{O}_k$ is the raw obstacle polygon set,
$\mathcal{O}_k^{+}$ is the inflated unsafe set, $\mathcal{C}_{\mathrm{safe},k}$
is the safe corridor, $\mathcal{G}_{\mathrm{goal},k}$ is the goal
region, and $\pi_k$ is a candidate path.

The function $\Phi(\cdot)$ encapsulates a multi-step geometric
pipeline. Image-space bounding polygons from the VLM are unprojected
to 3D ray directions using $K$ and $T_{\mathrm{cam}}$, then
intersected with the floor plane using $\mathcal{H}_{\mathrm{fp}}$
to produce 2D metric obstacle footprints in the robot workspace frame.
The safe corridor $\mathcal{C}_{\mathrm{safe},k}$ is derived from the
union of traversable-floor regions identified by the VLM, reduced by
the robot footprint to ensure the corridor represents center-point
reachable space. The goal region $\mathcal{G}_{\mathrm{goal},k}$ is
extracted from goal-related semantic cues such as open-door directions
or waypoint markers. This pipeline depends on camera calibration
accuracy; any systematic intrinsic or extrinsic calibration error
introduces a positional bias in the projected obstacle boundaries
that is not fully captured by the inflation model, making calibration
quality a practical limitation of the approach.

Obstacle inflation uses the Minkowski sum with a conservative safety
buffer $\mathcal{B}(r_{\mathrm{safe}})$:
\begin{equation}
\begin{aligned}
\mathcal{O}_k^{+} &= \mathcal{O}_k \oplus \mathcal{B}(r_{\mathrm{safe}}), \\
r_{\mathrm{safe}} &= r_{\mathrm{robot}} + r_{\mathrm{loc}}
+ r_{\mathrm{perc}} + r_{\mathrm{delay}} + r_{\mathrm{margin}},
\end{aligned}
\label{eq:inflation}
\end{equation}
where each term has a distinct physical origin. The footprint term
$r_{\mathrm{robot}}$ ensures the inflated region represents the
space the robot center must avoid to prevent physical contact with the
obstacle. The localization term $r_{\mathrm{loc}}$ accounts for
odometric drift and sensor noise in the robot's self-position estimate.
The perception term $r_{\mathrm{perc}}$ is a function of the VLM
confidence $\alpha_j$ and the estimated reprojection error: lower
confidence and higher reprojection error jointly increase $r_{\mathrm{perc}}$,
expanding the inflated region conservatively. The delay term
$r_{\mathrm{delay}}$ captures the position change the robot can
undergo during the full communication and processing cycle, ensuring
the safety check remains valid at command execution time rather than
only at observation time. The margin term $r_{\mathrm{margin}}$
provides a final conservative buffer against model errors and
unmodeled dynamics. Using a scalar inflation rather than an
anisotropic region is a deliberate simplification that sacrifices some
potential corridor width for robustness and computational tractability.
As illustrated in Figs.~\ref{fig:exp_uncertainty_3d}
and~\ref{fig:exp_confidence_safety_radius}, the appropriate buffer
cannot be determined from confidence alone; reprojection error must
also be considered.

A geometrically feasible candidate path satisfies
\begin{equation}
\begin{aligned}
\pi_k &= \{q_0,\ldots,q_N\}, \\
q_h &\in \mathcal{C}_{\mathrm{safe},k}, \quad
q_h \notin \mathcal{O}_k^{+}, \quad h = 0,\ldots,N.
\end{aligned}
\label{eq:path}
\end{equation}
Every candidate configuration must lie inside the safe corridor and
outside the inflated obstacle region. This geometric feasibility
condition is necessary but not sufficient for command approval: the
path must subsequently pass the full reachability gate before any
command is published.

\subsection{LLM Symbolic Task Reasoning}

Qwen3 via~Ollama reasons over the structured context $\mathcal{C}_k$,
comprising the validated scene representation and current robot states
of both platforms, to produce a symbolic task-level
proposal~\cite{kannan2024smartllm}:
\begin{equation}
\label{eq:llm_decision}
\begin{aligned}
d_k^{\mathrm{LLM}} &= \Psi(\mathcal{C}_k), \\
d_k^{\mathrm{LLM}} &\in \mathcal{D},
\end{aligned}
\end{equation}
The symbolic vocabulary covers the key behavioral modes of supervised
low-speed goal-directed navigation. PROCEED indicates that the scene
supports forward motion on the current candidate path. AVOID\_AND\_PROCEED
signals that an obstacle is present but an alternative traversable
corridor appears available. REPLAN indicates that the current path is
infeasible and that a new plan is required. HOLD is a conservative
pause while the system re-evaluates the scene. STOP is a terminal
directive for unsafe or emergency conditions. Qwen3 does not output
velocity values, steering angles, or any other continuous motor
command. Its output is a symbolic label that is treated as a candidate
input to the reachability gate, which computes the final verified
command:
\begin{equation}
u_{i,k}^{\mathrm{safe}} =
\Gamma\!\bigl(d_k^{\mathrm{LLM}},\;R_{i,k},\;
\mathcal{C}_{\mathrm{safe},k},\;\mathcal{O}_k^{+}\bigr).
\label{eq:llm_gate}
\end{equation}
The function $\Gamma(\cdot)$ is the formal safety filter between
task-level reasoning and physical execution. If the candidate control
derived from $d_k^{\mathrm{LLM}}$ violates any geometric or
reachability constraint, the gate rejects it regardless of the LLM
proposal. The LLM thus provides high-level structural guidance that
reduces the search space for candidate controls, but it exercises no
command authority.

\subsection{Robot Dynamics and Linearization}

Each platform is modeled for short-horizon reachability verification
using a reduced planar unicycle approximation:
\begin{equation}
\begin{aligned}
p_x(k{+}1) &= p_x(k) + \Delta t\, v(k)\cos\theta(k), \\
p_y(k{+}1) &= p_y(k) + \Delta t\, v(k)\sin\theta(k), \\
\theta(k{+}1) &= \theta(k) + \Delta t\, \omega(k),
\end{aligned}
\label{eq:unicycle}
\end{equation}
where $\Delta t$ is the sampling period and $\omega(k)$ is the
angular velocity. This model is justified for two reasons. First,
both the SVEA robot car and the Unitree~Go2 quadruped execute low-speed
planar motion in the tested laboratory environment: the SVEA performs
ground-plane navigation via Ackermann steering that is well-approximated
by a unicycle at low curvature, and the Go2 executes trot-gait
locomotion that reduces to planar dynamics at walking speed. Second,
the objective of the safety verification is to bound future planar
occupancy rather than to reproduce the full low-level mechanical
behavior of each platform; the short prediction horizon limits
error accumulation from the kinematic abstraction, and any
approximation error is treated as a bounded disturbance in the
verification. This model is therefore an intentional reduction for
safety envelope estimation, not a claim that both platforms share
identical dynamics.

Local linearization around nominal trajectories yields the affine
approximation
\begin{equation}
q_i(k{+}1) \approx
A_{i,k}\,q_i(k) + B_{i,k}\,u_i(k) + G_{i,k}\,w_i(k) + l_{i,k},
\label{eq:linear}
\end{equation}
where $A_{i,k} = \partial f / \partial q_i |_{q^*,u^*}$ is the
state-transition Jacobian evaluated at the nominal trajectory
$(q^*, u^*)$, $B_{i,k}$ is the input Jacobian, $G_{i,k}$ maps the
bounded disturbance $w_i(k)$ into the state space, and $l_{i,k}$ is
the affine linearization residual that bounds the approximation error
within the operating region. The residual $l_{i,k}$ is treated as
part of the disturbance set $\mathcal{W}_{i,k}$, ensuring the
propagated reachable tube accounts for the error introduced by
linearization. This affine structure enables efficient zonotope
propagation because affine maps applied to zonotopes produce
zonotopes in closed form.

\subsection{Zonotope Representation and Reachable Tube Propagation}

Reachable sets are represented as zonotopes to exploit their
closed-form propagation properties. A zonotope $\mathcal{Z}$ is
defined by a center $c$ and a generator matrix $G_Z$:
\begin{equation}
\mathcal{Z} = \langle c,\,G_Z\rangle
= \bigl\{c + G_Z\beta \mid \beta \in [-1,1]^m\bigr\},
\label{eq:zonotope}
\end{equation}
where $m$ is the number of generators. Each generator column in
$G_Z$ captures one independent uncertainty direction; the full
zonotope represents all points reachable by combinations of these
directions within unit $\ell_\infty$ bounds. This representation
is computationally effective because: (i)~the image of a zonotope
under an affine map is a zonotope, computable as
$A\mathcal{Z} = \langle Ac, AG_Z\rangle$; (ii)~the Minkowski sum
of two zonotopes is a zonotope obtained by concatenating their
generator matrices; and (iii)~the set-membership tests required for
the safety predicates reduce to linear feasibility problems over the
generators. As Minkowski sums accumulate generators, order-reduction
steps follow standard CORA practice to bound the generator count and
maintain tractable computation.

Given initial state set $R_{i,0} = \langle c_{i,0}, G_{i,0}\rangle$,
input set $\mathcal{U}_{i,k}$, and disturbance set $\mathcal{W}_{i,k}$,
the reachable tube is propagated recursively following
CORA-style set
propagation~\cite{althoff2021setpropagation,kochdumper2021sparse,liu2024refine}:
\begin{equation}
\begin{aligned}
R_{i,k+1} &=
A_{i,k}R_{i,k} \oplus B_{i,k}\mathcal{U}_{i,k}
\oplus G_{i,k}\mathcal{W}_{i,k} \oplus L_{i,k}, \\
R_i([0,T]) &= \bigcup_{k=0}^{N} R_{i,k}.
\end{aligned}
\label{eq:reach}
\end{equation}
The reachable tube $R_i([0,T])$ is the union of all step-wise
reachable sets over the prediction horizon $T$ and represents the
set of planar configurations that robot~$i$ may occupy under any
admissible input and bounded disturbance realization. The tube is
computed independently for Go2 and SVEA because the two robots
occupy different positions, headings, and kinematic envelopes at
every timestep. Even when the same shared scene model is used to
construct the geometric constraints, a command safe for Go2's
current reachable tube would not generally be safe for SVEA's, and
vice versa. A single averaged or shared command is therefore never
issued; each robot receives an independently verified directive.

\subsection{Safety Predicates and Decision Rule}

The centralized server evaluates three safety predicates against the
propagated reachable tube. The predicates are defined as
\begin{equation}
\begin{aligned}
\eta_{i,k}^{\mathrm{obs}} &:
R_i([0,T]) \cap \mathcal{O}_k^{+} = \emptyset, \\
\eta_{i,k}^{\mathrm{cor}} &:
R_i([0,T]) \subseteq \mathcal{C}_{\mathrm{safe},k}, \\
\eta_{k}^{\mathrm{int}} &:
R_G([0,T]) \cap R_S([0,T]) = \emptyset,
\end{aligned}
\label{eq:safety_predicates}
\end{equation}
where $G$ and $S$ denote the Go2 and SVEA reachable tubes
respectively. The first predicate requires the robot's reachable tube
to have empty intersection with every inflated obstacle region; this
prevents the robot from occupying any point classified as physically
unsafe under the compound inflation model. The second predicate
requires the tube to be contained within the admissible safe corridor;
a robot whose predicted trajectory exits the corridor may not collide
with a detected obstacle but is nonetheless moving outside the
verified traversable region. The third predicate prevents predicted
inter-robot collision by requiring that the Go2 and SVEA reachable
tubes do not overlap over the same prediction horizon. All three
predicates must be satisfied simultaneously for a command to be
approved.

The final safety decision is
\begin{equation}
d_{i,k} =
\begin{cases}
\mathrm{SAFE}, &
\eta_{i,k}^{\mathrm{obs}} \land \eta_{i,k}^{\mathrm{cor}}
\land \eta_{k}^{\mathrm{int}}, \\[4pt]
\mathrm{REPLAN}, &
\eta_{i,k}^{\mathrm{obs}} \land \neg\eta_{i,k}^{\mathrm{cor}}
\land \eta_{k}^{\mathrm{int}}, \\[4pt]
\mathrm{STOP}, &
\neg\eta_{i,k}^{\mathrm{obs}} \lor \neg\eta_{k}^{\mathrm{int}}.
\end{cases}
\label{eq:decision}
\end{equation}
This decision rule expresses the conservative logic of the safety
gate. When all predicates are satisfied, the command is approved and
the robot is directed to PROCEED or AVOID+PROCEED depending on the
LLM proposal. When the reachable tube avoids obstacles but exits the
safe corridor, replanning is required: the robot should hold its
position while a corridor-consistent alternative is computed. When
the reachable tube intersects an inflated obstacle region or predicts
inter-robot collision, the robot is commanded to STOP. Note that
the decision labels PROCEED and AVOID+PROCEED are both instances of
the SAFE outcome; the distinction is encoded in the candidate control
$u_{i,k}$ derived from the LLM proposal.

\subsection{Command Publication and Freshness}

The server packages each approved command into a robot-specific MQTT
command packet:
\begin{equation}
m_{i,k}^{\mathrm{cmd}} =
\bigl\{u_{i,k}^{\mathrm{safe}},\; d_{i,k},\; \sigma_{i,k},\;
t_k,\; T_{\mathrm{valid}}\bigr\},
\label{eq:mqtt_cmd}
\end{equation}
where $u_{i,k}^{\mathrm{safe}}$ is the verified control input,
$d_{i,k}$ is the safety decision label, $\sigma_{i,k}$ is the
safety flag, $t_k$ is the command timestamp, and $T_{\mathrm{valid}}$
is the command validity horizon. The command is robot-specific:
each of Go2 and SVEA receives a separately computed packet reflecting
its own current state, kinodynamic envelope, and verified reachable
set. No averaged or shared command is issued.

Each robot executes a received command only if the freshness
condition
\begin{equation}
0 \leq t_{\mathrm{now}} - t_k \leq T_{\mathrm{valid}},
\quad \sigma_{i,k} = \mathrm{SAFE}
\label{eq:freshness}
\end{equation}
is satisfied at the time of reception. If the elapsed time since the
command was computed exceeds $T_{\mathrm{valid}}$, the packet is
discarded as stale. This condition is safety-critical because the
centralized server computes commands based on a snapshot of the
world at time $t_k$. If the snapshot is old, the scene may have
changed: an obstacle may have moved into the planned corridor, or
a robot may have drifted from the state used in the reachability
computation. The validity horizon $T_{\mathrm{valid}}$ should be
set to exceed the total pipeline latency $T_{\mathrm{total}}$ with
a margin, but it should be short enough to ensure the command
remains relevant to the current physical configuration. Low-level
motor stabilization and emergency response remain on the robot side;
the centralized server provides supervisory safety-filtered
task-level command updates.

% ============================================================
\section{Experimental Setup and Online Validation}
\label{sec:experimental_validation}
% ============================================================

The validation examines whether the complete online pipeline operates
coherently under controlled laboratory conditions. The goal was not
offline replay or isolated module testing, but to evaluate whether
synchronized state acquisition, live semantic interpretation,
semantic-to-geometric conversion, reachability verification, and
command publication could operate as a causally ordered online loop
in which each command decision is generated from the current
synchronized sensing packet and is either approved or rejected before
reaching the robot execution interfaces.

\subsection{Platform Configuration}

The Unitree~Go2 quadruped served as the perception-capable platform.
It provided the live camera stream used to construct the shared
semantic--geometric scene representation and simultaneously published
odometry, IMU, and robot-state messages over the MQTT broker. The
SVEA robot car operated as the camera-less mobile platform: it
generated no local visual observations, but published its own
odometry, IMU, and robot-state messages and received robot-specific
verified commands. SVEA was not treated as a passive follower: it
actively contributed synchronized state data to every decision cycle
and received independently verified commands parameterized by its
own current pose, admissible input set, and reachable set, not those
of the Go2. The operational objective was cooperative
goal-directed navigation in an indoor environment containing static
structural elements and a moving human obstacle. Two scenarios were
evaluated: a clear-path case with static boundaries only, and a
dynamic-obstacle case with a human actor entering the shared workspace.

\subsection{Implementation Stack}

The Go2 camera stream was acquired through a ROS~2 image topic and
processed frame-by-frame by an OpenCV node running on the centralized
server. OpenCV generated the runtime diagnostic overlays, including
traversable-floor masks, obstacle boundaries, inflated unsafe regions,
candidate paths, reachable-tube projections, goal markers, and
safety-status panels. Mosquitto MQTT was used exclusively as the
asynchronous M2M transport layer for robot telemetry, server status
messages, and approved command packets. It performed no perception,
planning, reasoning, verification, or control; all decision logic
was implemented in the centralized server. Execution was performed
through ROS~2 Action interfaces for goal-level commands and
\texttt{cmd\_vel} topics for validated short-horizon velocity commands.

Semantic scene interpretation was performed by Qwen2.5-VL via~Ollama.
For each admitted camera frame, the VLM produced a JSON-structured
output containing semantic labels, detected-object bounding regions,
risk annotations, traversable-region descriptions, goal candidates,
and confidence values. The server applied schema validation before
any VLM output was used geometrically; malformed, incomplete, or
internally inconsistent outputs were discarded and did not propagate
to the reachability stage. Symbolic task-level reasoning was
performed by Qwen3 via~Ollama, which consumed the validated scene
representation and robot-state context to produce a task-level
proposal. This proposal carried no command authority. Final command
approval was assigned exclusively to the custom Python zonotope
module, which implemented CORA-style set propagation and acted as
the supervisory safety gate.

Table~\ref{tab:system_components} summarizes the role, output, and
command authority of each pipeline component. The table makes
explicit the central architectural constraint: perception and language
reasoning provide structured evidence and symbolic guidance, while
physical command authorization is granted solely by the zonotope
reachability gate.

\begin{table*}[!t]
\caption{System components, pipeline roles, and command authority.}
\label{tab:system_components}
\centering
\footnotesize
\setlength{\tabcolsep}{4pt}
\begin{tabularx}{\textwidth}{@{}p{0.15\textwidth}L L p{0.17\textwidth}@{}}
\toprule
\textbf{Component} & \textbf{Pipeline Role} &
\textbf{Produced Information} & \textbf{Command Authority} \\
\midrule
Unitree Go2 &
Vision-capable robot; publishes camera frames, odometry, IMU,
and robot-state data over MQTT. &
ROS~2 image topics; odometry, IMU, and state messages. &
None; receives approved commands only. \\
\addlinespace
SVEA robot car &
Camera-less robot; publishes odometry, IMU, and state; executes
independently verified robot-specific commands. &
Odometry, IMU, and state messages; used for robot-specific
reachability evaluation. &
None; receives approved commands only. \\
\addlinespace
Mosquitto MQTT &
Asynchronous M2M transport for telemetry, status messages,
and approved command packets. &
Routed state messages, server status, and verified commands. &
None; transport layer only. \\
\addlinespace
Qwen2.5-VL via Ollama &
Semantic interpretation of Go2 camera frames. &
Schema-validated JSON scene output with labels, regions,
and confidence values. &
None; perception evidence only. \\
\addlinespace
Qwen3 via Ollama &
Symbolic task-level reasoning from validated scene context
and robot states. &
Symbolic task proposal for downstream reachability verification. &
None; symbolic guidance only. \\
\addlinespace
Python zonotope module &
CORA-style set propagation; robot-specific predicate evaluation. &
Safety flag $\sigma_{k,i}$ and approved command
$u_{k,i}^{\mathrm{safe}}$. &
Sole command-approval authority. \\
\addlinespace
ROS~2 interfaces &
Forward verified commands to robot-side execution topics. &
Goal-level tasks and validated \texttt{cmd\_vel} messages. &
Forwarding after gate approval only. \\
\bottomrule
\end{tabularx}
\end{table*}

\subsection{Online Decision Loop}

At each runtime cycle, the server received the current Go2 camera
frame together with Go2 and SVEA state packets. Before semantic
interpretation or reasoning was initiated, the server enforced
timestamp consistency across all required streams using
Eq.~\eqref{eq:sync}. Only synchronized packets were admitted to the
VLM stage. The validated VLM output was converted into metric
geometric entities obstacle sets, inflated unsafe regions,
traversable regions, safe corridors, goal candidates, and candidate
paths consistent with Eq.~\eqref{eq:geometry}. The structured
scene context and robot states were then passed to Qwen3 to obtain
a symbolic task proposal. This proposal served only as an input
candidate for verification. The reachability module computed
independent reachable sets for Go2 and SVEA because the two robots
have different poses, kinematic envelopes, and workspace occupancy
at every timestep. For robot $i \in \{\mathrm{Go2},\,\mathrm{SVEA}\}$,
the command packet issued at timestep~$k$ is
\begin{equation}
m_{k,i}^{\mathrm{cmd}} =
\bigl\{u_{k,i}^{\mathrm{safe}},\; d_{k,i},\; \sigma_{k,i},\;
t_k,\; T_{\mathrm{valid}}\bigr\},
\quad i \in \{\mathrm{Go2},\,\mathrm{SVEA}\}.
\label{eq:multi_robot_command}
\end{equation}
No averaged or shared command was issued to both robots; each
command reflected the safety evaluation specific to that robot's
current state and reachable occupancy.

Algorithm~\ref{alg:online_loop} presents the full online decision
loop. The algorithm is structured as a strict sequence of gates; each
gate must pass before the next stage is evaluated. Failures in
synchronization, schema validation, geometric conversion,
reachability verification, or command freshness all result in
conservative rejection. This gated structure ensures that
language-model outputs cannot bypass geometric verification and that
no command reaches the execution interface without explicit
reachability approval.

\begin{algorithm*}[!t]
\footnotesize
\caption{Online centralized safety decision cycle.}
\label{alg:online_loop}
\begin{algorithmic}[1]
\Require Go2 frame $F_k$; states
$\mathbf{s}_k^{\mathrm{Go2}}$, $\mathbf{s}_k^{\mathrm{SVEA}}$
\Ensure Verified $m_{k,i}^{\mathrm{cmd}}$ or conservative directive
for each $i \in \{\mathrm{Go2},\mathrm{SVEA}\}$

\State Receive $F_k$, $\mathbf{s}_k^{\mathrm{Go2}}$,
$\mathbf{s}_k^{\mathrm{SVEA}}$ via MQTT transport.
\If{Eq.~\eqref{eq:sync} violated for any stream pair}
    \State Publish \textit{HOLD}/\textit{STOP}; log sync failure.
    \State \textbf{continue}
\EndIf
\State Query Qwen2.5-VL with $F_k$; obtain JSON output $\mathcal{J}_k$.
\If{$\mathcal{J}_k$ fails schema validation}
    \State Publish \textit{HOLD}; log schema failure.
    \State \textbf{continue}
\EndIf
\State Convert $\mathcal{J}_k$ to metric sets
$\{\mathcal{O}_k,\,\mathcal{O}_k^{+},\,\mathcal{C}_{\mathrm{safe},k},\,
\mathcal{G}_k,\,\pi_k\}$
via Eq.~\eqref{eq:geometry}.
\If{conversion incomplete or geometrically inconsistent}
    \State Publish \textit{HOLD}; log conversion failure.
    \State \textbf{continue}
\EndIf
\State Query Qwen3 with validated context; obtain proposal
$d_k^{\mathrm{LLM}}$.
\Statex \hspace{\algorithmicindent}%
\textit{Proposal is symbolic guidance only; no command authority.}
\For{each $i \in \{\mathrm{Go2},\mathrm{SVEA}\}$}
    \State Extract $x_k^i$, $\mathcal{U}^i$ from $\mathbf{s}_k^i$.
    \State Construct candidate $u_{k,i}$ from $d_k^{\mathrm{LLM}}$
    subject to $\mathcal{U}^i$.
    \State Propagate $R_i([0,T])$ via
    Eqs.~\eqref{eq:linear}--\eqref{eq:reach}.
    \If{$R_i([0,T]) \cap \mathcal{O}_k^{+} \neq \emptyset$}
        \State $d_{k,i} \gets \textit{STOP/HOLD}$;
        $\sigma_{k,i} \gets \textsc{Unsafe}$.
    \ElsIf{inter-robot reachable-set overlap detected}
        \State $d_{k,i} \gets \textit{HOLD}$;
        $\sigma_{k,i} \gets \textsc{Unsafe}$.
    \ElsIf{$R_i([0,T]) \not\subseteq \mathcal{C}_{\mathrm{safe},k}$}
        \State $d_{k,i} \gets \textit{REPLAN}$;
        $\sigma_{k,i} \gets \textsc{Unsafe}$.
    \Else
        \State $u_{k,i}^{\mathrm{safe}} \gets u_{k,i}$;
        $d_{k,i} \gets \textit{PROCEED/AVOID+PROCEED}$;
        $\sigma_{k,i} \gets \textsc{Safe}$.
    \EndIf
    \If{$\sigma_{k,i} = \textsc{Safe}$}
        \State Compose $m_{k,i}^{\mathrm{cmd}}$
        per Eq.~\eqref{eq:multi_robot_command}.
        \If{$T_{\mathrm{total}} > T_{\mathrm{valid}}$}
            \State Discard; publish \textit{HOLD};
            log stale-command event.
        \Else
            \State Publish $m_{k,i}^{\mathrm{cmd}}$
            to robot-$i$ execution topic.
        \EndIf
    \Else
        \State Publish conservative directive $d_{k,i}$
        to robot~$i$.
    \EndIf
\EndFor
\State Log frame ID, timestamps, states, $\mathcal{J}_k$,
geometric sets, $d_k^{\mathrm{LLM}}$, reachable sets, decisions,
commands, per-stage latencies, and any failure events.
\end{algorithmic}
\end{algorithm*}

\subsection{Experimental Scenarios and Safety Precautions}

Two experimental scenarios were evaluated. In the clear-path scenario,
the workspace contained only static structural boundaries and an
open-door goal region; no dynamic obstacle was present. In the
dynamic-obstacle scenario, a human actor was introduced as a moving
obstacle in the shared workspace. All trials were performed under
supervised laboratory conditions. The robots operated under speed
limits and within predefined workspace boundaries with emergency-stop
access. No Go2 or SVEA velocity command was forwarded to a robot
execution interface unless it had passed the reachability gate. These
measures constitute responsible experimental precautions and should
not be interpreted as formal safety certification or as a guarantee
of safe behavior outside the tested conditions.

\subsection{Experimental Figures}

\begin{figure*}[h]
\centering
\includegraphics[width=\textwidth,keepaspectratio]{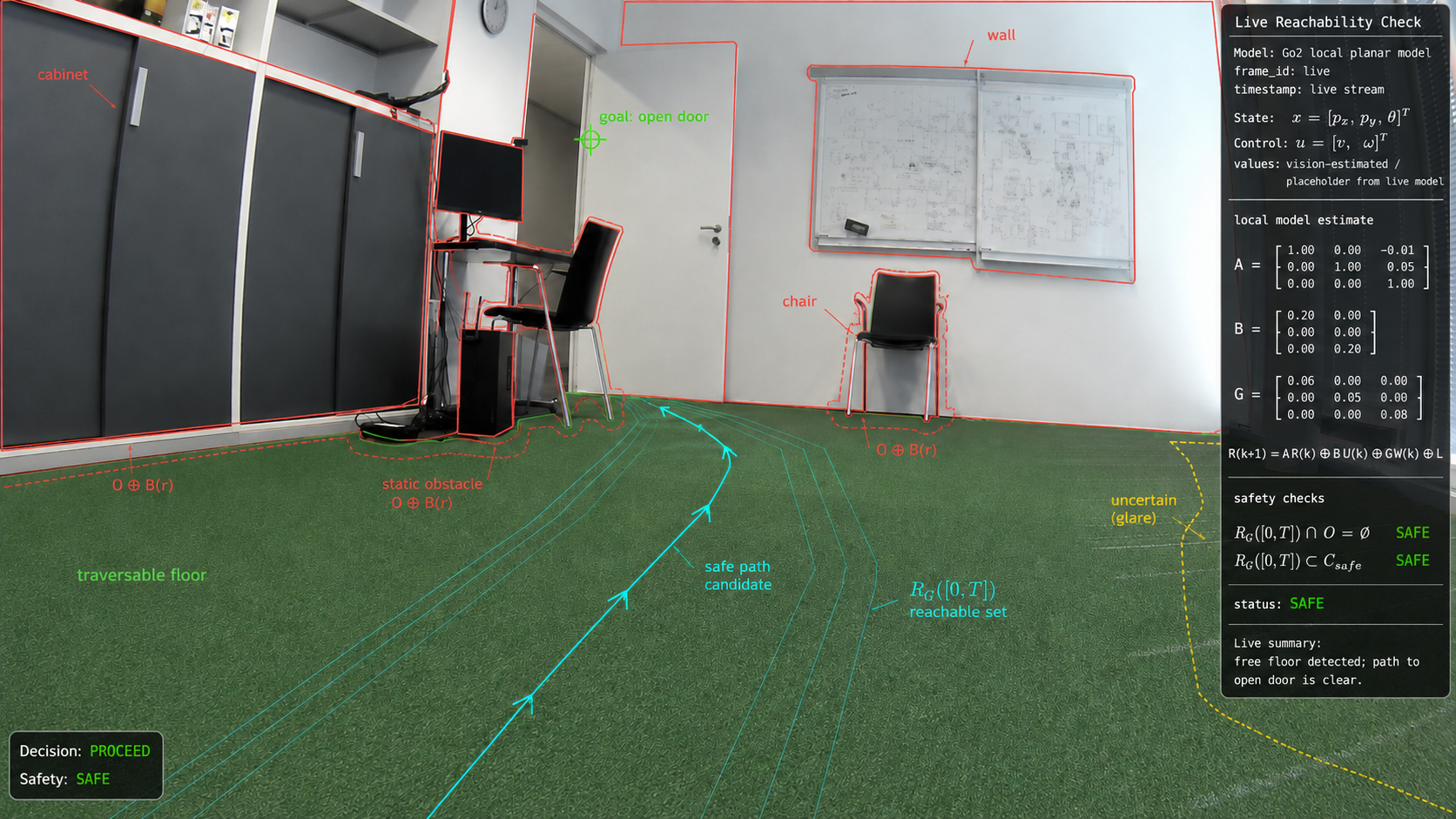}
\caption{Online clear-path validation from the Go2 camera stream.}
\label{fig:exp_clear_path_overlay}
\end{figure*}

Figure~\ref{fig:exp_clear_path_overlay} shows the nominal clear-path validation
case from the Go2 camera stream. The overlay identifies the traversable floor,
static obstacle boundaries, the open-door goal region, the candidate path, and
the corresponding reachable tube. The detected cabinet, wall, and chairs are
treated as static constraints, whereas the visible floor area is mapped to the
admissible safe corridor $\mathcal{C}_{\mathrm{safe},k}$. The candidate path is generated toward the open-door goal and is then verified using the local
reachability gate. In the displayed case, the reachable tube $R_G([0,T])$ remains fully contained
within the admissible corridor and does not intersect the inflated obstacle set
$\mathcal{O}_{k}^{+}$. Consequently, the predicates
$R_G([0,T]) \cap \mathcal{O}_{k}^{+}=\emptyset$ and
$R_G([0,T]) \subseteq \mathcal{C}_{\mathrm{safe},k}$ are both satisfied, and
the server assigns the decision \textit{PROCEED}. This figure illustrates that
motion approval is not triggered by visual recognition alone, but only after the camera derived semantic scene representation is converted into conservative geometric constraints and accepted by the reachability-based safety gate.

\begin{figure}[h]
\centering
\includegraphics[width=\columnwidth,keepaspectratio]%
{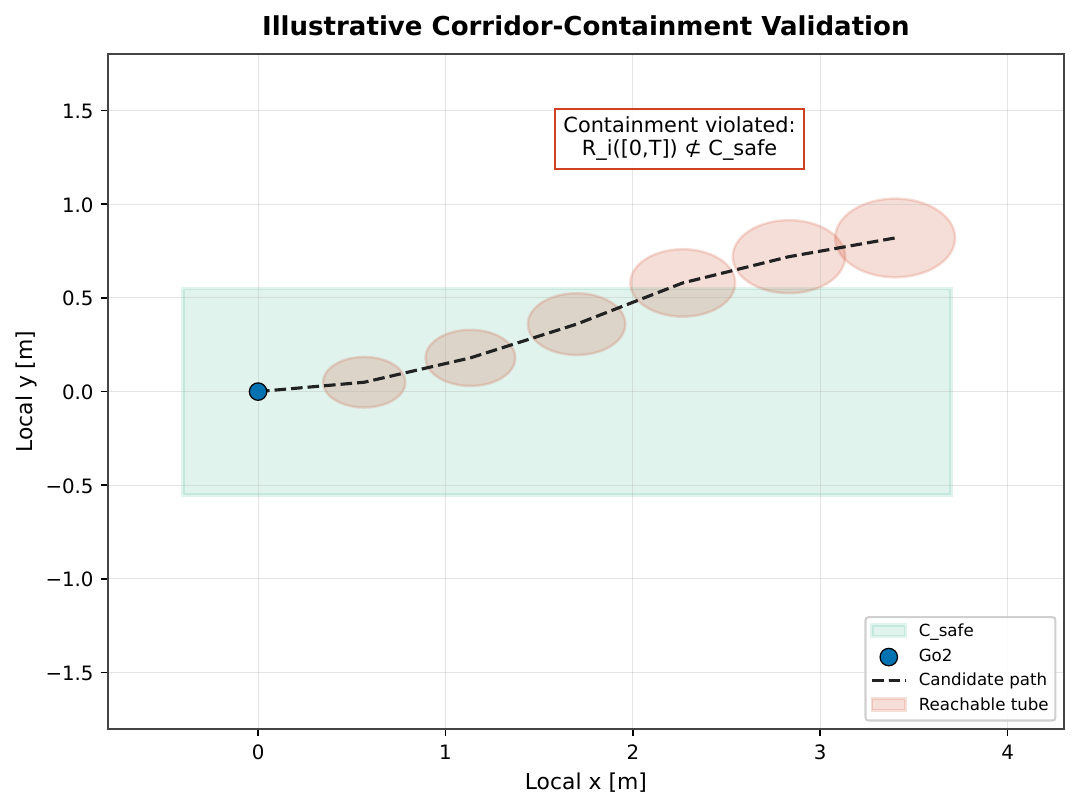}
\caption{Corridor-containment in the local workspace.}
\label{fig:exp_corridor_containment}
\end{figure}

Figure~\ref{fig:exp_corridor_containment} illustrates the corridor-containment
validation case in the local robot-centered workspace. The plot shows the
admissible safe corridor $\mathcal{C}_{\mathrm{safe}}$, the current Go2 pose,
the candidate path, and the propagated reachable tube over the finite horizon
$[0,T]$. Although the candidate motion may remain separated from the detected
obstacle set, the reachable tube partially extends outside the verified
traversable corridor. This results in a violation of the containment condition
$R_i([0,T]) \subseteq \mathcal{C}_{\mathrm{safe}}$.

This example isolates the second safety predicate in
Eq.~\eqref{eq:safety_predicates}. It demonstrates that obstacle clearance alone
is not sufficient for command approval. A motion can be collision-free with
respect to the inflated obstacle set while still being unsafe if it leads the
robot toward an unverified, occluded, or non-traversable region outside the
admissible corridor. Therefore, the reachability gate rejects the candidate
command and assigns the supervisory decision \textit{REPLAN}. The command is
withheld until a new candidate path is generated whose reachable tube remains
fully contained inside $\mathcal{C}_{\mathrm{safe}}$ while also satisfying the
obstacle-separation predicate.

\begin{figure*}[h]
\centering
\includegraphics[width=\textwidth,keepaspectratio]{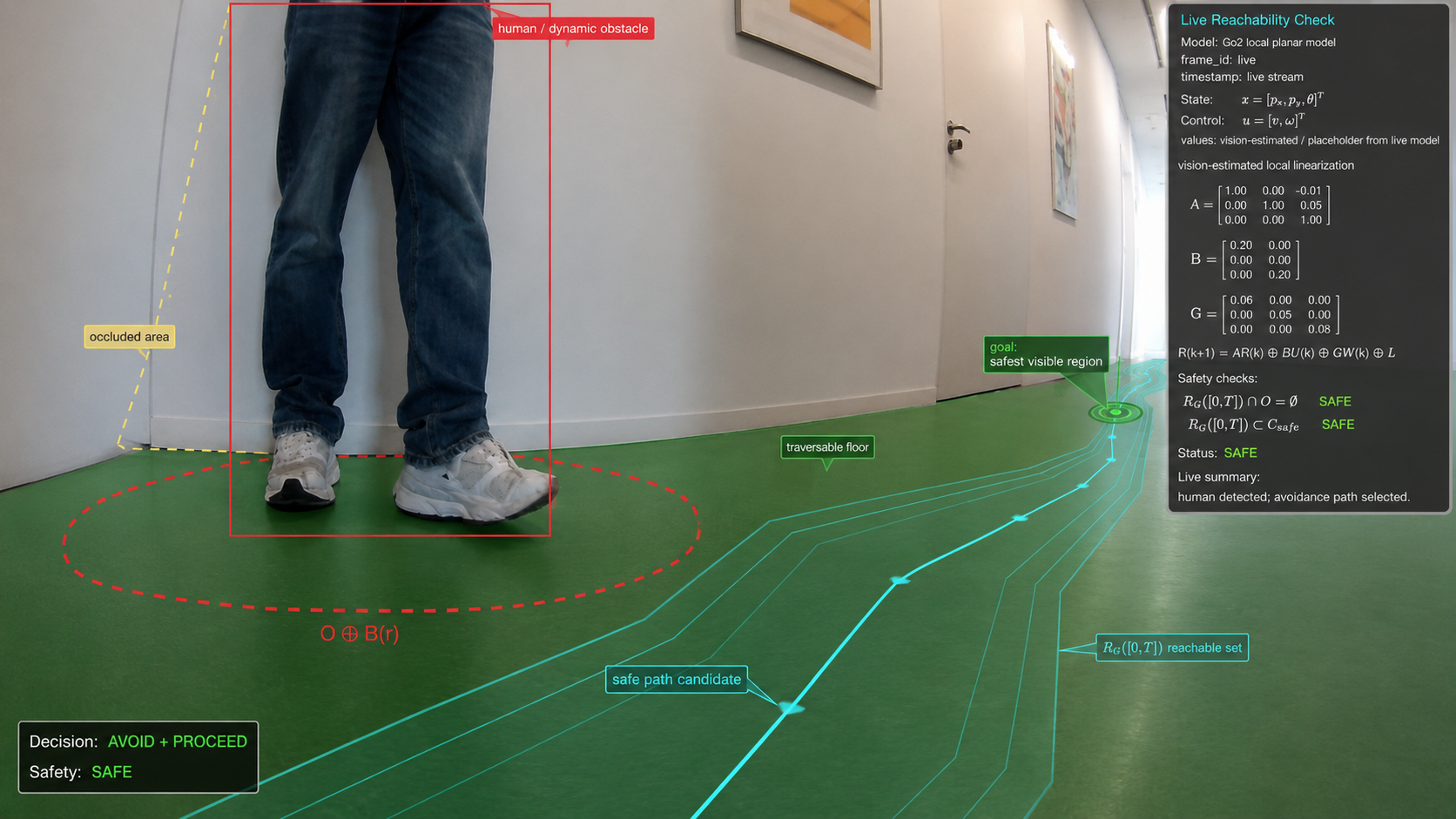}
\caption{Online dynamic-obstacle validation from the Go2 camera stream.}
\label{fig:exp_dynamic_obstacle_overlay}
\end{figure*}

Figure~\ref{fig:exp_dynamic_obstacle_overlay} presents the online
dynamic-obstacle validation case using a live frame from the Go2 camera
stream. The detected human actor is interpreted as a dynamic obstacle and is
therefore not treated only as a semantic object, but as a safety-critical
region that directly affects motion approval. The red detection box represents
the VLM-based semantic localization of the human, while the dashed red region
indicates the conservatively inflated unsafe set
$\mathcal{O}_{k}^{+}$ obtained from the obstacle footprint using
Eq.~\eqref{eq:inflation}. This inflation accounts for perception uncertainty,
ground-plane projection error, and the motion uncertainty associated with the
dynamic obstacle.

The figure also illustrates how the semantic output is converted into an
actionable geometric constraint for planning. The traversable floor region is
identified as the candidate free space, whereas the occluded area is marked as
uncertain and excluded from direct motion preference. Instead of selecting a
straight path toward the goal, the planner redirects the candidate trajectory
through the remaining visible and traversable corridor. The cyan path therefore
represents an avoidance-aware candidate path, and the surrounding cyan tube
represents the predicted reachable set $R_G([0,T])$ of the Go2 over the finite
verification horizon. The goal is assigned to the safest visible region rather
than to an unconstrained image-space target, ensuring that the selected motion
remains compatible with the perceived scene geometry.

The reachability status panel on the right reports the online validation
process used before command publication. The local planar model, state vector,
control vector, and uncertainty terms are evaluated to propagate the reachable
set. The motion command is accepted only when the reachable tube satisfies the
two safety predicates: separation from the inflated obstacle region and
containment within the safe corridor. In the displayed case, the conditions
$R_G([0,T]) \cap \mathcal{O}_{k}^{+}=\emptyset$ and
$R_G([0,T]) \subset \mathcal{C}_{\mathrm{safe},k}$ are both satisfied, resulting
in the supervisory decision \textit{AVOID+PROCEED}. This demonstrates that the
VLM detection alone does not authorize robot motion. Instead, the semantic
interpretation is first transformed into conservative geometric constraints,
and the final command is approved only after explicit reachability-gate
validation confirms that the redirected motion remains safe.

\begin{figure}[h]
\centering
\includegraphics[width=\columnwidth,keepaspectratio]%
{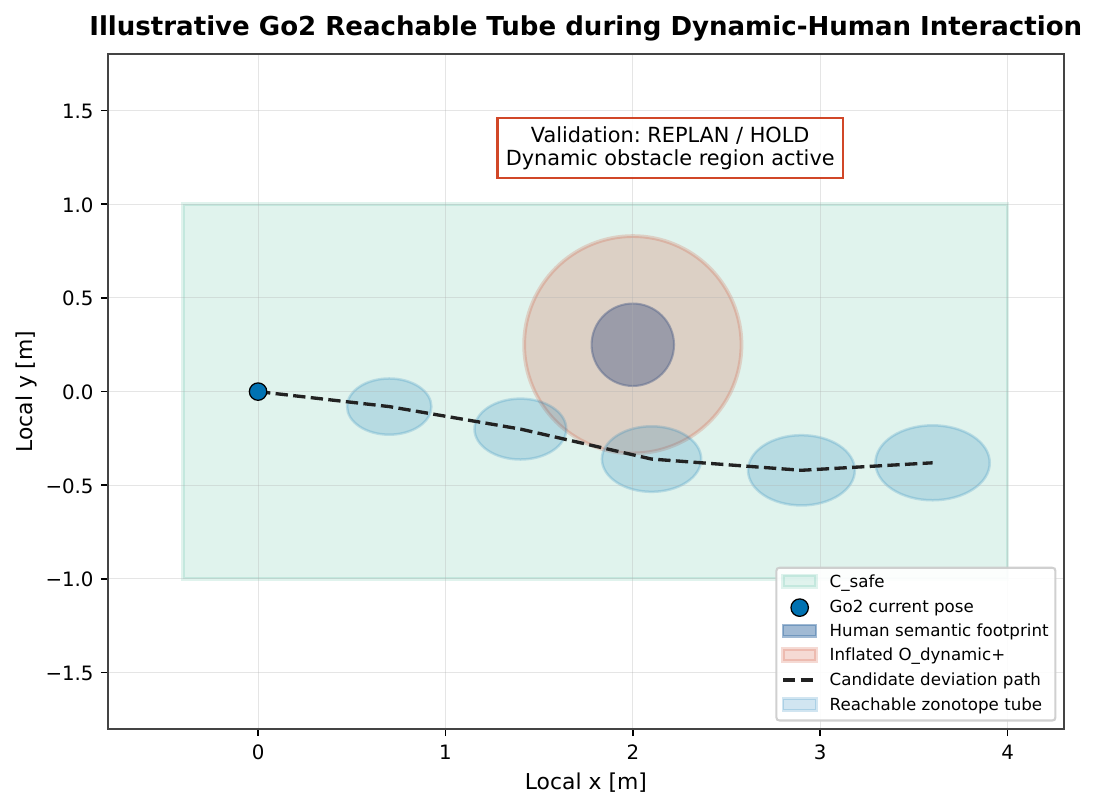}
\caption{Reachability under dynamic-obstacle interaction.}
\label{fig:exp_dynamic_human_reachability}
\end{figure}

Figure~\ref{fig:exp_dynamic_human_reachability} shows a
dynamic-obstacle case. The plot includes the admissible
corridor, human semantic footprint, inflated unsafe region, candidate
deviation path, and propagated zonotope reachable tube. When the
reachable tube intersects the inflated obstacle region or fails
corridor containment, the server issues \textit{REPLAN} or
\textit{HOLD}. This figure demonstrates that the safety gate
evaluates the geometry of the proposed motion, not merely the
presence or absence of a detected obstacle: a redirected path that
clears the human footprint but exits the safe corridor is still
rejected and a new clean path has been provided to be executed.

\begin{figure}[h]
\centering
\includegraphics[width=\columnwidth,keepaspectratio]%
{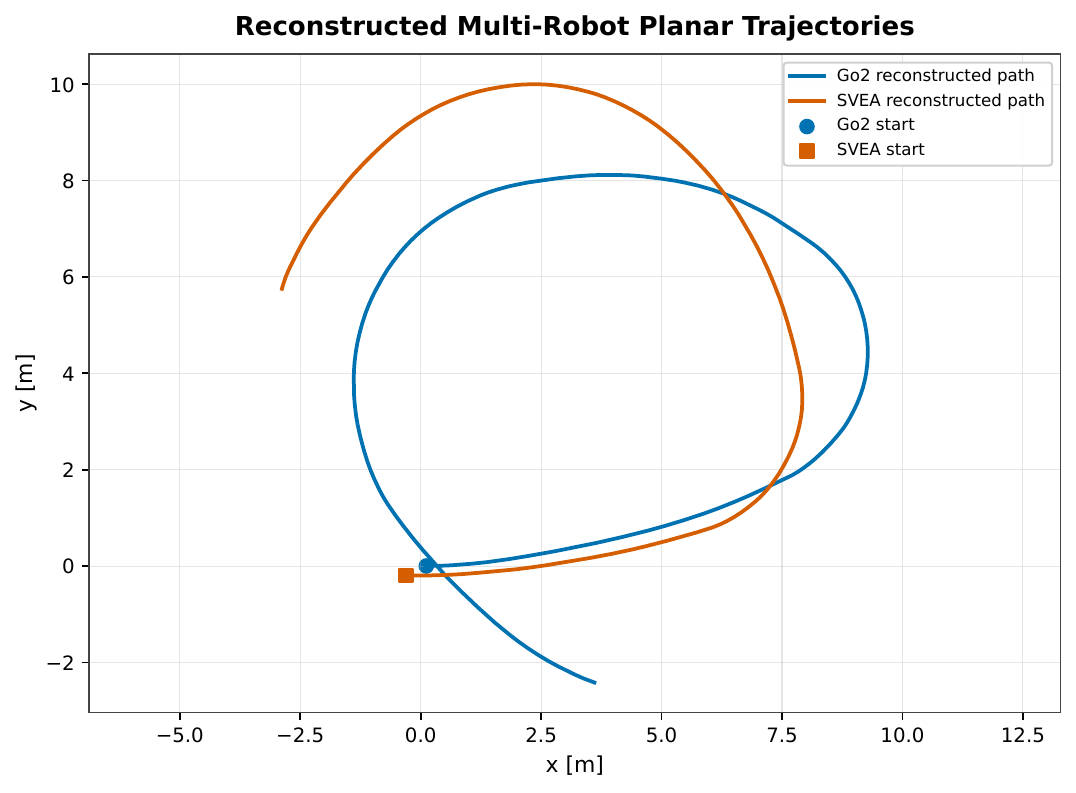}
\caption{Reconstructed Go2 and SVEA planar trajectories.}
\label{fig:M2M4}
\end{figure}

Figure~\ref{fig:M2M4} shows the reconstructed planar trajectories
of Go2 and SVEA in the global $x$--$y$ workspace. The Go2 trajectory
reflects the motion of the vision-capable robot supplying the live
visual stream. The SVEA trajectory reflects the motion of the
camera-less robot executing server-approved commands derived from
Go2-shared perception and verified against SVEA's own synchronized
state. Separated traces with distinct initial conditions confirm
that both robots actively participated in the cooperative task and
received independently verified robot-specific commands. The
separation of the trajectories also illustrates why a single shared
command would be architecturally incorrect: the two robots occupy
different positions in the workspace at every timestep and must
have their reachable sets checked independently against the same
geometric constraints.

\begin{figure}[h]
\centering
\includegraphics[width=\columnwidth,keepaspectratio]%
{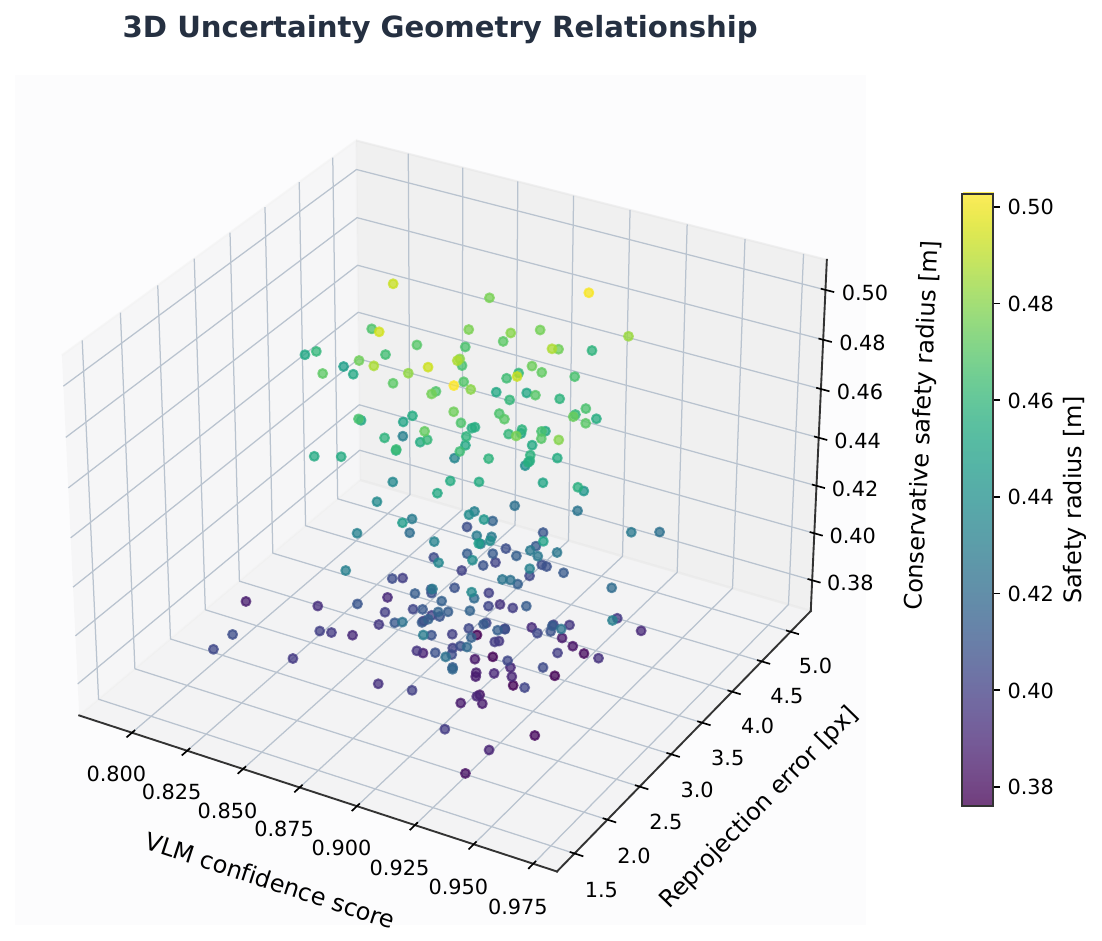}
\caption{Confidence, reprojection error, and safety-radius surface.}
\label{fig:exp_uncertainty_3d}
\end{figure}
Figure~\ref{fig:exp_uncertainty_3d} visualizes the coupled influence of
semantic confidence and geometric projection uncertainty on the conservative
safety radius assigned to detected obstacles. The horizontal axes represent the VLM confidence score $\alpha_j$ and the ground-plane reprojection error, while the vertical axis and color scale indicate the resulting safety radius used to
inflate the obstacle footprint. The figure highlights that semantic reliability
alone is not sufficient for safe motion planning. A detection can receive a high VLM confidence score because the object category is recognized correctly, while its projected ground-plane location may still be uncertain due to camera perspective, shallow viewing angles, imperfect calibration, partial occlusion, or object-height ambiguity.

This distinction is important for the proposed safety-aware pipeline because
the robot does not only require a correct semantic label; it also requires a
geometrically reliable obstacle boundary. As the reprojection error increases,
the corresponding safety radius remains conservatively enlarged, even when the
semantic confidence is relatively high. Conversely, when the reprojection error is low, the obstacle can be represented with a smaller inflation margin, provided that the semantic confidence remains sufficient. Therefore, the figure supports the use of a compound uncertainty-aware inflation rule in
Eq.~\eqref{eq:inflation}, where both semantic uncertainty and geometric
projection error contribute to the final inflated obstacle set
$\mathcal{O}_{k}^{+}$. This prevents the planner from over-trusting visually
confident but geometrically uncertain detections and ensures that the reachable set is evaluated against conservative obstacle boundaries before any motion command is accepted.
\begin{figure}[h]
\centering
\includegraphics[width=\columnwidth,keepaspectratio]%
{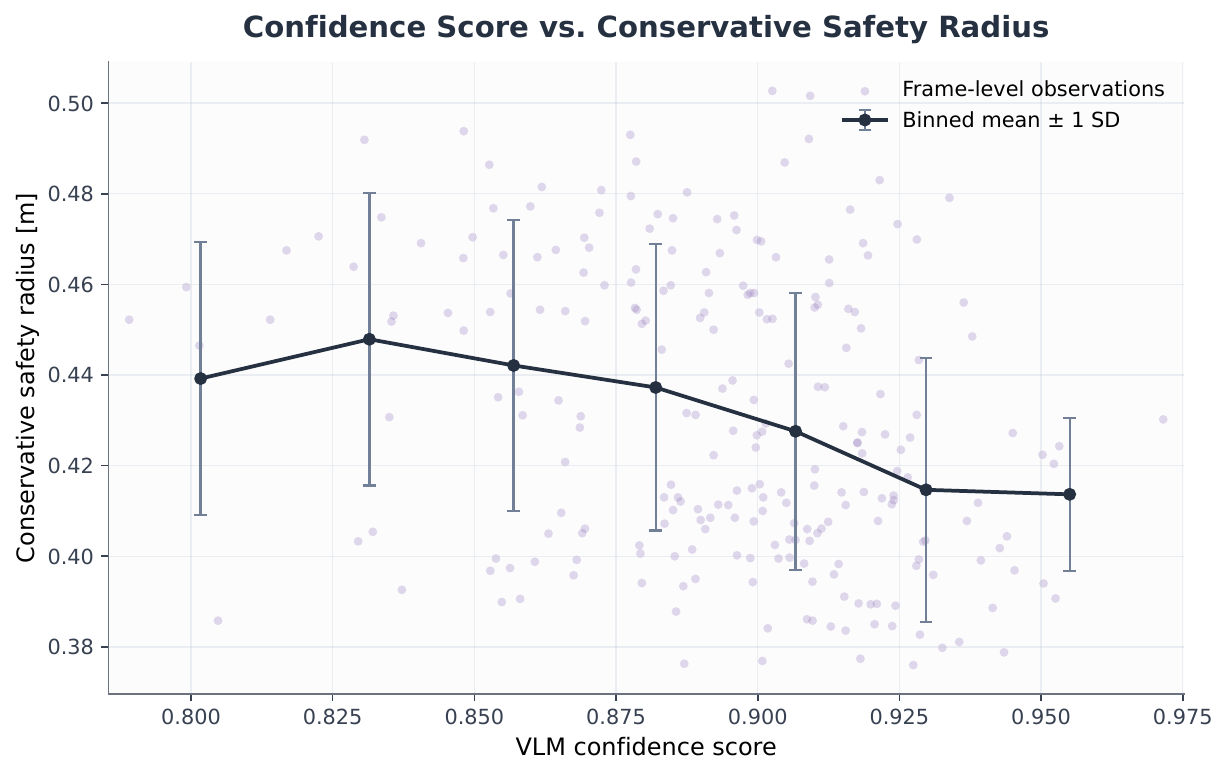}
\caption{VLM confidence versus conservative safety radius.}
\label{fig:exp_confidence_safety_radius}
\end{figure}

Figure~\ref{fig:exp_confidence_safety_radius} provides a
confidence-centered view of the safety-radius behavior. The plotted
trend supports the expected relationship that lower confidence
generally requires more conservative inflation. However, the
variability across confidence levels confirms that confidence alone
is an insufficient determinant of the spatial safety buffer.
Reprojection error, calibration quality, and visual ambiguity can
all require a larger radius even at high confidence, supporting the
multi-source inflation model.
However, the figure also shows that the relationship is not purely monotonic or deterministic. The visible spread of the frame-level samples and the overlapping standard-deviation intervals demonstrate that VLM confidence alone is not sufficient to determine the final safety radius. Even at relatively high confidence levels, the radius may remain enlarged when the corresponding obstacle projection is affected by ground-plane reprojection error, calibration uncertainty, partial visibility, perspective distortion, or ambiguous object boundaries. Therefore, the plotted distribution supports the use of a multi-source inflation strategy rather than a confidence-only rule. In the proposed framework, the confidence score contributes to the semantic reliability assessment, while geometric uncertainty determines how conservatively the obstacle footprint must be inflated before reachability verification is performed. This result is particularly important for the safety gate because the reachable set is not evaluated against the raw detected obstacle boundary, but against the inflated unsafe set $\mathcal{O}_{k}^{+}$. By preserving a larger radius when
semantic or geometric uncertainty remains significant, the planner avoids
over-trusting visually confident detections and maintains a conservative
separation margin before issuing any motion command.

\subsection{Reproducibility}

Each run logged the following data to enable reconstruction of every
online decision cycle from stored inputs and intermediate outputs:
frame identifiers; camera, odometry, IMU, and state timestamps for
both Go2 and SVEA; MQTT send and server-receive timestamps; VLM JSON
output; schema-validation result; LLM prompt and response; obstacle
sets; inflated unsafe regions; safe corridors; goal regions; candidate
paths; reachability parameters; reachable tubes; decision labels;
safety flags; command packets; command validity horizons; per-stage
latency values; failure events; and full configuration files. The
configuration record included robot initial states and goal
definitions, obstacle settings, safety-buffer parameters, prediction
horizon, MQTT communication settings, model identifiers for
Qwen2.5-VL and Qwen3, and any stochastic seeds used by the
experimental scripts. This logging structure ensures that each
decision cycle can be reconstructed offline for analysis and
verification.

% ============================================================
\section{Results and Discussion}
\label{sec:results}
% ============================================================

The online validation produced outcomes consistent with the expected
behavior of the pipeline under the tested laboratory conditions.
This section discusses the key observations across the two scenarios
and the rejection events observed at each pipeline gate.

\subsection{Test Matrix}
\label{subsec:test_matrix}
Table~\ref{tab:test_matrix} lists the evaluation scenarios, the objective of
each, and the safety predicate each is designed to exercise. The scenarios were
chosen so that every predicate is stressed in isolation at least once, and so
that the nominal-acceptance case (S1) and the maximal-rejection case (S2) bound
gate behavior from both sides. All scenarios were recorded in the CPS-TUM
Multi-Robot Systems laboratory at $30$\,Hz, giving $1{,}953$ consecutive
decision cycles over $65.1$\,s of operation. At each cycle the arc-rollout
planner submitted $9$ curvature candidates to the reachability engine, so the
statistics below summarize $1{,}953 \times 9 = 17{,}577$ verified candidate
arcs, each propagated over $N = 15$ zonotope steps at $\Delta t = 0.1$\,s
($T = 1.5$\,s) with disturbance envelope
$\mathcal{W} = \mathrm{diag}(0.06,\,0.05,\,0.08)$.

\begin{table*}[t]
\caption{Test matrix: evaluation scenarios, objectives, and the safety
predicate each is designed to exercise.}
\label{tab:test_matrix}
\centering
\footnotesize
\setlength{\tabcolsep}{4pt}
\begin{tabularx}{\textwidth}{@{}c p{0.16\textwidth} c c L p{0.13\textwidth}@{}}
\toprule
\textbf{ID} & \textbf{Scenario} & \textbf{Cycles} & \textbf{Duration (s)} &
\textbf{Objective} & \textbf{Predicate stressed} \\
\midrule
S1 & Office room, goal approach & 720 & 24.0 &
Establish nominal acceptance: verify the gate approves motion when a
traversable corridor and a goal cue are both present and no dynamic obstacle
is in the workspace. &
$\eta^{\mathrm{obs}}$, $\eta^{\mathrm{cor}}$ \\
\addlinespace
S2 & Stairwell, non-traversable hazard & 361 & 12.0 &
Establish maximal rejection: verify that a region which is geometrically open
but not traversable is excluded by corridor containment even though no obstacle
intersection occurs. &
$\eta^{\mathrm{cor}}$ (isolated) \\
\addlinespace
S3 & Corridor, pedestrian entry & 421 & 14.0 &
Verify dynamic-obstacle handling: compound inflation of a moving human
footprint and redirection of the candidate path through the remaining
traversable corridor. &
$\eta^{\mathrm{obs}}$ \\
\addlinespace
S4 & Sustained human tracking & 451 & 15.0 &
Verify sustained operation under continuous re-planning at close range,
including decision stability and clearance maintenance. &
$\eta^{\mathrm{obs}}$, $\eta^{\mathrm{cor}}$ \\
\midrule
\multicolumn{2}{@{}l}{\textbf{Total}} & \textbf{1{,}953} & \textbf{65.1} &
\multicolumn{2}{l}{$17{,}577$ verified candidate arcs, $N = 15$ steps each} \\
\bottomrule
\end{tabularx}
\end{table*}

\subsection{Clear-Path Scenario}

In the clear-path scenario, the VLM produced a structured scene
description identifying a traversable floor region, static corridor
walls, and an open-door goal cue. After schema validation, the
semantic-to-geometric conversion module constructed a conservative
safe corridor and an obstacle set consisting only of the static
structural boundaries. The candidate path satisfied the geometric
feasibility condition in Eq.~\eqref{eq:path}: it lay within the
safe corridor and outside the inflated obstacle set. The reachable
tubes for both Go2 and SVEA, propagated over the finite prediction
horizon, satisfied all three predicates in
Eq.~\eqref{eq:safety_predicates}: obstacle avoidance, corridor
containment, and inter-robot separation. The reachability gate
accordingly assigned \textit{PROCEED} and published approved
robot-specific velocity commands to each platform. This outcome
validates the nominal acceptance behavior of the pipeline: when
the shared scene representation supports geometrically feasible
navigation and the reachable tube confirms it under bounded
disturbances, the gate correctly approves motion without unnecessary
conservatism.

\subsection{Dynamic-Obstacle Scenario}

When the human actor entered the workspace, the VLM produced a
bounding region with an associated confidence score. The geometric
conversion translated this detection into an inflated unsafe region
parameterized by the full compound inflation model in
Eq.~\eqref{eq:inflation}. The initially proposed candidate path
intersected the inflated region. The LLM symbolic proposal shifted
to \textit{AVOID+PROCEED}, directing the candidate path through the
remaining traversable corridor. The reachability gate evaluated the
redirected candidate: when the redirected reachable tube avoided the
inflated obstacle region, remained inside the safe corridor, and did
not predict inter-robot collision, the gate approved the command.
When the redirected tube failed any predicate, the gate issued
\textit{REPLAN} or \textit{HOLD}.

These outcomes confirm the correct operation of the layered safety
architecture. The VLM accurately detected the human presence and
provided the geometric inputs required for safe-region construction;
however, the command was not issued on the basis of VLM detection
alone. The LLM directed the system toward an avoidance strategy, but
the physical command was withheld until the zonotope reachability
gate confirmed geometric feasibility. The two language models thus
served their intended advisory roles without exercising command
authority.

\subsection{Quantitative Gate Behavior}
\label{subsec:gate_behavior}
Table~\ref{tab:quant_results} reports, per scenario, the fraction of cycles in
which the gate approved motion, the fraction in which it withheld motion, the
distribution of the internal safety state, and the mean cycle risk. Three
observations follow.

First, the gate is not uniformly conservative: it approves motion in $56.4\%$
of cycles in the clear-path scenario S1 and in $0.0\%$ of cycles at the
stairwell in S2, a $56.4$ percentage-point spread driven entirely by scene
geometry. A filter that behaved identically in both cases would be either
useless or paralyzing. Second, the internal safety state tracks the scenario in
the expected direction: \textsc{safe} occupancy falls monotonically from
$92.2\%$ (S1) to $18.8\%$ (S4) as obstacle proximity increases, and the
\textsc{unsafe} state, absent in S1 and S2, reaches $68.5\%$ in S4 where a
pedestrian is tracked at sub-metre range. Third, S2 isolates the practical
value of the corridor predicate: the stairwell produces no object to intersect,
so $\eta^{\mathrm{obs}}$ holds in all $361$ cycles and only
$\eta^{\mathrm{cor}}$ rejects the motion. A pipeline verifying obstacle
intersection alone would have approved forward motion toward a descending
staircase in $100\%$ of those cycles and all $3{,}249$ candidate arcs.

\begin{table*}[!t]
\caption{Gate behavior per evaluation scenario. Approval and withholding are
reported as fractions of decision cycles; the internal safety state is the
fraction of cycles spent in each condition; risk is the mean cycle-level score
$\pm$ one standard deviation.}
\label{tab:quant_results}
\centering
\footnotesize
\setlength{\tabcolsep}{4pt}
\renewcommand{\arraystretch}{1.15}
\begin{tabular*}{\textwidth}{@{\extracolsep{\fill}} c l rr rr rrr c @{}}
\toprule
& & \multicolumn{2}{c}{\textbf{Volume}}
  & \multicolumn{2}{c}{\textbf{Gate outcome (\%)}}
  & \multicolumn{3}{c}{\textbf{Internal safety state (\%)}} & \\
\cmidrule(lr){3-4} \cmidrule(lr){5-6} \cmidrule(lr){7-9}
\textbf{ID} & \textbf{Scenario} &
\textbf{Cycles} & \textbf{Arcs} &
\textbf{Approved} & \textbf{Withheld} &
\textbf{Safe} & \textbf{Warning} & \textbf{Unsafe} &
\textbf{Mean risk} \\
\midrule
S1 & Office, goal approach  & 720 & 6{,}480 & 56.4 & 43.6  & 92.2 & 7.8  & 0.0  & $0.187 \pm 0.092$ \\
S2 & Stairwell hazard       & 361 & 3{,}249 & 0.0  & 100.0 & 48.5 & 51.5 & 0.0  & $0.341 \pm 0.126$ \\
S3 & Corridor, pedestrian   & 421 & 3{,}789 & 47.7 & 52.3  & 78.9 & 17.1 & 4.0  & $0.272 \pm 0.186$ \\
S4 & Sustained tracking     & 451 & 4{,}059 & 37.9 & 62.1  & 18.8 & 12.6 & 68.5 & $0.678 \pm 0.230$ \\
\midrule
\multicolumn{2}{@{}l}{\textbf{All scenarios}}
& \textbf{1{,}953} & \textbf{17{,}577}
& \textbf{39.8} & \textbf{60.2}
& \textbf{64.3} & \textbf{19.0} & \textbf{16.7}
& $\mathbf{0.347 \pm 0.248}$ \\
\bottomrule
\end{tabular*}
\end{table*}

\subsection{Decision-Conditioned Statistics}
\label{subsec:decision_stats}
Table~\ref{tab:decision_stats} conditions risk and clearance on the decision
actually issued. The pattern is the strongest available evidence that the gate
responds to geometry rather than to semantics. \textsc{proceed} is issued at
the lowest mean risk ($0.155$) and the largest mean clearance ($2.43$\,m);
\textsc{avoid+proceed} is issued at the \emph{highest} mean risk of any class
($0.737$) and the smallest clearance ($0.92$\,m), which is the intended
behavior since that decision is reserved for cases where an obstacle is close
but a verified corridor still exists. The withholding decisions occupy the
intermediate band ($0.336$ and $0.410$).

Decision stability is quantified by the transition rate. Over $65.1$\,s the
gate changed its issued decision $31$ times, a switching rate of $0.476$\,Hz
against a $30$\,Hz cycle rate; the decision was held constant across $98.4\%$
of consecutive cycle pairs. This matters because a filter oscillating between
approve and withhold at the cycle rate would produce unusable actuation.

\begin{table}[t]
\caption{Statistics conditioned on the decision issued by the gate. Clearance
is the mean distance to the closest detected obstacle in cycles carrying that
decision.}
\label{tab:decision_stats}
\centering
\footnotesize
\setlength{\tabcolsep}{3pt}
\begin{tabularx}{\columnwidth}{@{}L c c c c@{}}
\toprule
\textbf{Decision} & \textbf{Cycles} & \textbf{Share} &
\textbf{Mean risk} & \textbf{Clearance (m)} \\
\midrule
\multicolumn{5}{@{}l}{\textit{Motion approved}} \\
PROCEED           & 247 & $12.6\%$ & $0.155$ & $2.43$ \\
AVOID $+$ PROCEED & 201 & $10.3\%$ & $0.737$ & $0.92$ \\
TURN LEFT         & 170 & $8.7\%$  & $0.300$ & $1.70$ \\
TURN RIGHT        & 160 & $8.2\%$  & $0.167$ & $1.56$ \\
\addlinespace
\multicolumn{5}{@{}l}{\textit{Motion withheld}} \\
WAIT              & 914 & $46.8\%$ & $0.336$ & $1.53$ \\
CAUTION           & 261 & $13.4\%$ & $0.410$ & n/a \\
\midrule
\textbf{Total}    & \textbf{1{,}953} & $100\%$ & $0.347$ & --- \\
\bottomrule
\end{tabularx}
\end{table}

\subsection{Perception Statistics and Validity of the Uncertainty Model}
\label{subsec:stat_uncertainty}
Table~\ref{tab:perception_stats} summarizes the $973$ detections that entered
geometric conversion. Two properties directly motivate the compound inflation
rule of Eq.~\eqref{eq:inflation}.

First, confidence is class-dependent and is \emph{not} aligned with hazard.
Persons were detected with the highest mean confidence ($0.870 \pm 0.068$) and
at the closest range ($0.91 \pm 0.26$\,m, minimum $0.51$\,m), while cabinets at
comparable range ($1.00 \pm 0.42$\,m) were detected with the lowest confidence
($0.613 \pm 0.086$). A confidence-proportional margin would assign the
\emph{smallest} buffer to the closest and most safety-critical class.
A further $89$ detections ($9.1\%$) carried no resolvable confidence; these
were retained and treated as maximum-perception-uncertainty cases rather than
discarded, since discarding them would remove a real obstacle from
$\mathcal{O}_k^{+}$.

Second, the correlation structure is decisive. Across the $794$ cycles carrying
at least one confidence-bearing detection, the Pearson correlation between the
minimum per-cycle detection confidence and the cycle-level risk is
$r = 0.301$ ($r^{2} = 0.091$). Across the $855$ cycles carrying at least one
detection, the correlation between minimum obstacle distance and cycle risk is
$r = -0.607$ ($r^{2} = 0.368$). Semantic confidence therefore explains $9.1\%$
of the variation in cycle risk while obstacle geometry explains $36.8\%$ a
factor of $4.1$. A confidence-only inflation rule would mis-specify the safety
buffer in the large majority of cycles. This is the quantitative justification
for decomposing $r_{\mathrm{safe}}$ rather than scaling a single term by
$\alpha_j$, and it converts
Figs.~\ref{fig:exp_uncertainty_3d} and~\ref{fig:exp_confidence_safety_radius}
from illustrations into evidence.

The detection load was modest: a mean of $0.50$ objects per cycle with a
maximum of $2$, and $56.2\%$ of cycles carrying no detection. Cycles without
detections are not treated as free space; the corridor predicate still applies,
which is why S2 rejects every candidate despite an empty obstacle set.

\begin{table}[t]
\caption{Perception statistics over the $973$ detections admitted to geometric
conversion.}
\label{tab:perception_stats}
\centering
\footnotesize
\setlength{\tabcolsep}{3pt}
\begin{tabularx}{\columnwidth}{@{}L c c c c@{}}
\toprule
\textbf{Class} & \textbf{Detections} & \textbf{Confidence} &
\textbf{Range (m)} & \textbf{Min (m)} \\
\midrule
Person   & 456 & $0.870 \pm 0.068$ & $0.91 \pm 0.26$ & $0.51$ \\
Chair    & 404 & $0.842 \pm 0.083$ & $3.19 \pm 2.94$ & $0.99$ \\
Cabinet  & 113 & $0.613 \pm 0.086$ & $1.00 \pm 0.42$ & $0.64$ \\
\midrule
All      & 973 & $0.844 \pm 0.095$ & $1.86 \pm 2.21$ & $0.51$ \\
\multicolumn{5}{@{}l}{\footnotesize Unresolved confidence: 89 ($9.1\%$),
treated as maximum uncertainty.} \\
\multicolumn{5}{@{}l}{\footnotesize Dynamic: 456 ($46.9\%$);
static: 517 ($53.1\%$).} \\
\bottomrule
\end{tabularx}
\end{table}

The measured motion envelope also supports the reduced unicycle abstraction of
Eq.~\eqref{eq:unicycle}: mean forward speed was $0.091$\,m\,s$^{-1}$ with a
peak of $1.007$\,m\,s$^{-1}$, and mean angular rate $|\omega|$ was
$0.054$\,rad\,s$^{-1}$ with a peak of $0.650$\,rad\,s$^{-1}$. At these rates
the per-step displacement over $\Delta t = 0.1$\,s is at most $0.10$\,m with a
heading change of at most $0.065$\,rad, so the linearization residual
$l_{i,k}$ absorbed into $\mathcal{W}$ stays small relative to the fixed
disturbance envelope.

\subsection{Rejection Cases and Conservative Behavior}

Rejection events were observed when: the reachable tube intersected
the inflated obstacle set (\textit{STOP/HOLD}); the reachable tube
exited the safe corridor (\textit{REPLAN}); the inter-robot separation
predicate was violated (\textit{HOLD}); a synchronization failure
was detected (\textit{HOLD/STOP}); or the VLM produced a
schema-inconsistent output (\textit{HOLD}). In every rejection case,
the safety gate overrode the LLM symbolic proposal, and no command
reached the execution interface. This conservative gating behavior
is an intended property of the design: it is preferable to command
both robots to hold their positions while the system re-evaluates
than to issue a command based on inconsistent or geometrically unsafe
data.

\subsection{Multi-Robot Independence and SVEA Active Role}

Go2 and SVEA received different robot-specific commands at every
timestep because they occupied different positions, headings, and
velocity states. The command issued to Go2 was not shared with SVEA;
each command reflected the independent safety evaluation for that
robot. The reconstructed trajectories in Fig.~\ref{fig:M2M4} confirm
that both robots participated actively in the cooperative task.
SVEA's contributions were not passive: it supplied synchronized
odometry, IMU, and state data at every cycle, and the reachability
predicate for SVEA was evaluated using SVEA's own kinematic state
and reachable set, not an approximation derived from the Go2 state.
This independence is essential to the correctness of the safety
evaluation: two robots at different positions can have overlapping
or conflicting reachable tubes even when each individually avoids
the static obstacle set.

\subsection{Comparison with the State of the Art}
\label{subsec:sota_discussion}
The capability comparison in Table~\ref{tab:sota_comparison} can be sharpened
using the measured results, in three concrete contrasts.

Against language-model planners without a runtime gate
(\cite{ahn2023saycan,driess2023palme,zitkovich2023rt2,liang2023codeaspolicies,huang2023innermonologue,shah2023lmnav}),
the operative difference is the $60.2\%$ of cycles in which motion was
withheld. In each of those cycles the LLM had already produced a symbolic
proposal and the gate overrode it; since none of these systems exposes a
kinodynamic verification stage, the corresponding proposals would have been
executed. The $0.0\%$ approval rate at the stairwell (S2) is the sharpest
instance: the semantic layer correctly identified traversable floor, and only
the corridor predicate prevented motion toward a descending staircase.

Against reachability planners without semantic input
(\cite{kousik2020rtd,liu2024refine,michaux2024sparrows}), the operative
difference is that those methods require an externally supplied obstacle set.
In S2 there is no obstacle to supply: the hazard is a semantic property
(non-traversability), not a geometric one. Supplying
$\mathcal{C}_{\mathrm{safe},k}$ from semantic evidence is what allows a
set-based method to reject that case at all.

Against the single-robot LLM-plus-reachability filter~\cite{hafez2025safe}, the
operative difference is constraint transfer and the inter-robot predicate. The
single-robot formulation offers no mechanism by which a robot without a camera
obtains $\mathcal{O}_k^{+}$ and $\mathcal{C}_{\mathrm{safe},k}$, and no
predicate coupling two tubes; neither is a parameter change.

A direct numerical benchmark on a common task was not attempted, and we regard
reporting one as inadvisable: the baselines differ in platform, sensing
modality and success criterion, and none publishes a command-approval rate for
a camera-less robot verified against a partner's perception. The defensible
comparison is the capability matrix together with the measured behavior of the
proposed gate.

\subsection{Generalizability to Real-World Deployment}
\label{subsec:generalizability}
The results support a bounded rather than a general claim, and it is worth
being precise about where the boundary lies.

\textit{What is expected to transfer.} The separation between advisory
semantics and verified command authority is platform-independent: it requires
only a bounded input set and a disturbance envelope per robot, which any mobile
platform admits. The five-gate cascade with conservative fallback carries over
unchanged. The finding that semantic confidence is a weak predictor of
geometric risk ($r^{2} = 0.091$ against $0.368$) is a property of monocular
ground-plane projection rather than of this laboratory, and should be expected
wherever obstacle geometry is recovered from a single camera.

\textit{What is expected to degrade.} The approval rates in
Table~\ref{tab:quant_results} are scenario-conditioned and are not a
performance figure for the method. Denser environments will lower them: the
mean detection load here was $0.50$ objects per cycle, and a scene with ten
simultaneous obstacles would both enlarge $\mathcal{O}_k^{+}$ and shrink
$\mathcal{C}_{\mathrm{safe},k}$, plausibly driving approval toward the S2
regime unless the inflation terms are re-tuned. The $1.5$\,s verification
horizon is adequate at $0.09$\,m\,s$^{-1}$ mean speed but must grow with
velocity, and tube volume grows with horizon, so conservatism and speed trade
against each other directly.

\textit{What does not transfer without further work.} Outdoor operation
invalidates the flat-ground assumption underlying
$\mathcal{H}_{\mathrm{fp}}$; variable lighting degrades the semantic stage in
ways the inflation model cannot detect; and the centralized server remains a
single point of failure that per-robot verification does not compensate for.
These are stated in Section~\ref{sec:limitations} and form the substance of the
future work in Section~\ref{sec:conclusion}. The appropriate reading is that
this paper establishes the mechanism and characterizes it quantitatively in a
controlled setting, not that it certifies the mechanism for deployment.

\subsection{Decision Logic Summary}

Table~\ref{tab:decision_logic} summarizes the mapping between
reachability conditions, active predicates, assigned decisions, and
command status. The table is consistent with the decision rule in
Eq.~\eqref{eq:decision} and reflects the operational behavior
observed during the online validation. Approved commands
(PROCEED and AVOID+PROCEED) were only issued when all three
reachability predicates were satisfied and the command freshness
condition in Eq.~\eqref{eq:freshness} was met.

\begin{table*}[h]
\caption{Reachability gate decision logic.}
\label{tab:decision_logic}
\centering
\footnotesize
\setlength{\tabcolsep}{4pt}
\begin{tabularx}{\textwidth}{@{}L p{0.24\textwidth} c c@{}}
\toprule
\textbf{Condition} & \textbf{Active Predicate(s)} &
\textbf{Decision} & \textbf{Status} \\
\midrule
All predicates satisfied &
$\eta^{\mathrm{obs}} \land \eta^{\mathrm{cor}} \land \eta^{\mathrm{int}}$ &
PROCEED & Approved \\
\addlinespace
Obstacle present; alt.\ path available &
$\eta^{\mathrm{obs}} \land \eta^{\mathrm{cor}} \land \eta^{\mathrm{int}}$
(after redirect) &
AVOID+PROCEED & Approved \\
\addlinespace
Obstacle-free; corridor violated &
$\eta^{\mathrm{obs}} \land \neg\eta^{\mathrm{cor}} \land \eta^{\mathrm{int}}$ &
REPLAN & Withheld \\
\addlinespace
Obstacle intersection &
$\neg\eta^{\mathrm{obs}}$ &
STOP/HOLD & Withheld \\
\addlinespace
Inter-robot overlap &
$\neg\eta^{\mathrm{int}}$ &
HOLD & Withheld \\
\addlinespace
Synchronization failure &
Gate~1 (Eq.~\eqref{eq:sync}) &
HOLD/STOP & Withheld \\
\addlinespace
VLM schema failure &
Gate~2 &
HOLD & Withheld \\
\addlinespace
Stale command &
$T_{\mathrm{total}} > T_{\mathrm{valid}}$ &
HOLD & Withheld \\
\bottomrule
\end{tabularx}
\end{table*}

% ============================================================
\section{Latency and Runtime Analysis}
\label{sec:latency}
% ============================================================

The end-to-end decision cycle latency determines how frequently the
centralized server can deliver fresh safety-verified commands to the
robot fleet and directly governs the choice of the command validity
horizon $T_{\mathrm{valid}}$. The total pipeline latency is
decomposed as
\begin{align}
T_{\mathrm{total}} =\;
& T_{\mathrm{MQTT}}
+ T_{\mathrm{sync}}
+ T_{\mathrm{VLM}}
+ T_{\mathrm{geo}}
\nonumber\\
&+ T_{\mathrm{LLM}}
+ T_{\mathrm{reach}}
+ T_{\mathrm{cmd}},
\label{eq:latency}
\end{align}
where each term is the contribution of one pipeline stage.
Table~\ref{tab:latency_budget} provides a representative
design-time latency budget for the complete pipeline. \emph{The
values in Table~\ref{tab:latency_budget} are not reported as
instrumentally measured results}; they constitute a representative
online runtime budget derived from the known characteristics of the
hardware, software stack, and network configuration used in the
laboratory experiment. Instrumentally measured latency benchmarking
over extended trials is identified as future work.

\begin{table*}[h]
\caption{Representative per-cycle design-time latency budget.}
\label{tab:latency_budget}
\centering
\footnotesize
\setlength{\tabcolsep}{4pt}
\begin{tabularx}{\textwidth}{@{}p{0.22\textwidth} c c c L@{}}
\toprule
\textbf{Pipeline Stage} & \textbf{Symbol} &
\textbf{Range (ms)} & \textbf{Budget (ms)} & \textbf{Rationale} \\
\midrule
MQTT transport &
$T_{\mathrm{MQTT}}$ & 5--25 & 15 &
Local LAN or loopback broker; single-hop, low contention. \\
\addlinespace
Synchronization \& validation &
$T_{\mathrm{sync}}$ & 1--8 & 5 &
Timestamp-pair checking, alignment, duplicate and
out-of-order rejection. \\
\addlinespace
VLM inference \& JSON validation &
$T_{\mathrm{VLM}}$ & 350--1200 & 750 &
Qwen2.5-VL via Ollama; range depends on GPU availability,
image resolution, and prompt complexity. \\
\addlinespace
Semantic-to-geometric conversion &
$T_{\mathrm{geo}}$ & 5--35 & 20 &
Homography projection, Minkowski inflation, corridor
and goal-region construction. \\
\addlinespace
LLM symbolic reasoning &
$T_{\mathrm{LLM}}$ & 120--600 & 300 &
Qwen3 via Ollama processing compact JSON scene context
and robot states. \\
\addlinespace
Reachability propagation (${\times}2$) &
$T_{\mathrm{reach}}$ & 20--160 & 100 &
Short-horizon zonotope propagation and three-predicate
evaluation per robot; $50\,\mathrm{ms}{\times}2$. \\
\addlinespace
Command serialization \& publication &
$T_{\mathrm{cmd}}$ & 2--15 & 10 &
Safe-command packet formation, MQTT publication,
and ROS~2 forwarding. \\
\midrule
\textbf{Total} &
$T_{\mathrm{total}}$ & --- & \textbf{1200} &
Representative online design-time budget. \\
\bottomrule
\end{tabularx}
\end{table*}

The dominant latency contributor is VLM inference
($T_{\mathrm{VLM}} = 750\,\mathrm{ms}$), which depends strongly on
the availability of local GPU acceleration, the resolution of the
input camera frame, and the complexity of the structured output schema.
LLM symbolic reasoning ($T_{\mathrm{LLM}} = 300\,\mathrm{ms}$) is
the second largest contributor; its range is narrower than VLM
inference because the input to Qwen3 is a compact JSON context
rather than a raw image. Reachability propagation
($T_{\mathrm{reach}} = 100\,\mathrm{ms}$ for two robots) scales
linearly with the number of robots; adding a third robot would
increase this term by approximately $50\,\mathrm{ms}$ under the
same hardware configuration. MQTT transport, synchronization,
geometric conversion, and command publication collectively contribute
approximately $50\,\mathrm{ms}$ under laboratory network conditions.

The adopted total budget of $T_{\mathrm{total}} \approx 1200\,\mathrm{ms}$
implies that the command validity horizon must satisfy
\begin{equation}
T_{\mathrm{valid}} \geq T_{\mathrm{total}} + \delta_{\mathrm{margin}},
\label{eq:validity_horizon}
\end{equation}
where $\delta_{\mathrm{margin}}$ accounts for variability in inference
time and network jitter. In the experiments, $T_{\mathrm{valid}}$ was
set conservatively to $1.5\,\mathrm{s}$ in the clear path scenario
and $2.0\,\mathrm{s}$ in the dynamic obstacle scenario, reflecting
the higher scene variability in the latter case. A command that
arrived outside $T_{\mathrm{valid}}$ was rejected as stale even
if all reachability predicates had been satisfied. This rejection
is not a pipeline failure; it is a safety mechanism that prevents
commands computed from old perception data from being applied to a
physical scene that may have changed.

The adopted budget indicates that the proposed implementation is
appropriate for supervisory low speed local navigation and
safety-filtered command publication under indoor operation. It is
not designed to replace high frequency low level stabilization loops,
which operate at substantially higher rates on the robot side. The
centralized server provides safety filtered task level command updates
at the supervisory rate dictated by $T_{\mathrm{total}}$; on robot
controllers handle stabilization and emergency braking at their
native frequencies.

% ============================================================
\section{Baseline and Ablation Evaluation}
\label{sec:ablation}
% ============================================================

To isolate the contribution of each pipeline component, the validation
protocol defined baseline conditions that disable complete modules
and ablation conditions that vary selected parameters.
Table~\ref{tab:baselines_ablations} summarizes all conditions.
Because the evaluation was conducted under the tested laboratory
configuration and does not include quantitative logging from
instrumented module removal, the table presents the evaluation
protocol and the contribution each condition is intended to isolate,
rather than numerical performance differences. Systematic quantitative
evaluation across these conditions is identified as future work.

\begin{table*}[h]
\caption{Baseline and ablation evaluation conditions.}
\label{tab:baselines_ablations}
\centering
\footnotesize
\setlength{\tabcolsep}{4pt}
\begin{tabularx}{\textwidth}{@{}p{0.05\textwidth} p{0.22\textwidth} L L@{}}
\toprule
\textbf{ID} & \textbf{Condition} &
\textbf{Pipeline Modification} &
\textbf{Isolated Contribution} \\
\midrule
B1 & No reachability gate &
Reachability module disabled; LLM proposals published
without geometric verification. &
Role of the reachability gate in preventing unsafe
command execution. \\
\addlinespace
B2 & No VLM perception &
VLM removed; geometric conversion receives no semantic
scene evidence. &
Role of online semantic perception in constructing
obstacle sets, safe corridors, and goal regions. \\
\addlinespace
B3 & VLM$+$LLM without reachability &
Both language models retained; reachability gate removed. &
Whether language-model outputs alone suffice for safe
command publication. \\
\addlinespace
B4 & No shared perception for SVEA &
SVEA does not receive Go2-derived scene evidence and
relies only on its own state feedback. &
Role of shared perception for the camera-less robot's
safety constraint construction. \\
\addlinespace
A1 & Obstacle inflation sweep &
$r_{\mathrm{safe}}$ varied; remaining pipeline unchanged. &
Sensitivity of command approval rate to obstacle buffering. \\
\addlinespace
A2 & Perception uncertainty sweep &
$r_{\mathrm{perc}}$ varied with VLM confidence. &
Sensitivity to semantic uncertainty in the safety-margin
model. \\
\addlinespace
A3 & Communication delay sweep &
Synthetic MQTT delay injected prior to synchronization
and command publication. &
Sensitivity to timing jitter, synchronization rejection
rate, and stale command behavior. \\
\addlinespace
A4 & Reachability horizon sweep &
Prediction horizon $T$ varied; robot constraints fixed. &
Sensitivity of corridor compliance and command approval
to reachable set horizon length. \\
\bottomrule
\end{tabularx}
\end{table*}

The baseline conditions (B1--B4) remove complete pipeline modules.
Under B1, LLM proposals are published without geometric verification;
this is expected to produce commands that enter inflated obstacle
regions or exit the safe corridor in the dynamic-obstacle scenario,
isolating the specific contribution of the reachability gate. Under
B2, the absence of semantic perception prevents construction of
obstacle sets and safe corridors, eliminating the geometric inputs
to the reachability check. Under B3, both language models contribute
but the reachability gate is absent; this condition tests whether
VLM scene interpretation and LLM symbolic reasoning together are
sufficient to produce safe commands without a formal verification
layer, which the architecture hypothesis predicts they are not.
Under B4, SVEA cannot receive geometric scene constraints derived
from Go2's perception; it can only use its proprioceptive state,
which limits its ability to evaluate the obstacle and corridor
predicates.

The ablation conditions (A1--A4) vary single parameters while
preserving the full pipeline structure. A1 examines the
trade-off between conservative buffering and command acceptance
rate: a larger $r_{\mathrm{safe}}$ reduces the number of approved
commands but provides a wider safety margin. A2 tests how changes
in perceived semantic uncertainty propagate through the inflation
model to the reachability predicates. A3 introduces synthetic
communication delay to examine the synchronization rejection rate
and the effectiveness of the freshness condition in
Eq.~\eqref{eq:freshness}: as MQTT delay increases, more commands
are expected to arrive beyond $T_{\mathrm{valid}}$. A4 examines
how the prediction horizon length affects the size of the reachable
tube and the resulting corridor-containment rejection rate.

% ============================================================
\section{Limitations}
\label{sec:limitations}
% ============================================================

The following limitations define the operational scope of the
reported validation and should be considered when assessing the
generalizability of the proposed pipeline. They do not invalidate
the architecture, but they establish the conditions under which the
reported results apply.

The validation is restricted to two robots operating in controlled
indoor laboratory environments with static structural boundaries,
a single moving human obstacle, and relatively uncluttered scenes.
The pipeline has not been evaluated in outdoor environments, under
variable lighting, with occluded camera views, with multiple
simultaneous dynamic obstacles, or in deployments with more than
two robots. Broader validation across these conditions is required
before deployment claims beyond the tested setting can be supported.

The safety predicates depend critically on the quality of VLM scene
interpretation and camera calibration. Calibration errors introduce
systematic bias in the projected obstacle positions that the
inflation model cannot fully compensate. VLM failures, partial
detections, or significant domain shift between training and
deployment conditions can produce incorrect obstacle sets or safe
corridors. The compound inflation formulation in
Eq.~\eqref{eq:inflation} partially addresses perception uncertainty
through $r_{\mathrm{perc}}$, but it cannot recover from systematic
VLM failures or severely degraded visual conditions.

The centralized server is a potential bottleneck and single point
of failure. All perception interpretation, reasoning, and
reachability verification are concentrated in one process; a server
failure would interrupt the command pipeline for all robots.
Extension to decentralized or fault-tolerant architectures is
identified as future work.

VLM and LLM inference latency limits the command update rate to
the supervisory timescale analyzed in Section~\ref{sec:latency}.
This rate is appropriate for low-speed local navigation but would
be insufficient for high-speed or highly dynamic tasks. Faster
inference through hardware acceleration, asynchronous pipelines,
or smaller domain-specific models could reduce this limitation.

The reduced unicycle model used for reachability propagation
captures planar occupancy over a short finite horizon. It does not
represent the full nonholonomic kinematics of the SVEA platform
or the legged locomotion dynamics of the Go2 under all operating
conditions. The model is appropriate for the safety envelope
estimation task as formulated, but a more complete dynamic model
would reduce conservatism and improve command acceptance rates in
more demanding maneuvers.

The proposed pipeline is not a formally safety-certified controller.
It provides a safety-filtering mechanism that demonstrates, under
the tested laboratory conditions, that reachability-gated command
publication supports safer coordination than unfiltered
language-model-based control. Broader safety claims require
systematic validation across diverse scenarios, formal worst-case
analysis, hardware-in-the-loop testing, and long-duration operational
trials. Actual measured latency benchmarking over extended runs is
required if the manuscript is to report measured runtime performance
rather than design-time estimates.

% ============================================================
\section{Conclusion}
\label{sec:conclusion}
% ============================================================

This paper addressed the problem of safe coordination in heterogeneous
M2M robotic systems where platforms differ structurally in sensing
capability. The central challenge is that the camera-less SVEA robot
car cannot observe its environment independently, yet must navigate
safely in coordination with the vision-capable Unitree~Go2 quadruped.
The proposed pipeline resolves this asymmetry by enabling the Go2 to
supply shared semantic--geometric scene awareness to SVEA through an
MQTT broker and a centralized server, while SVEA contributes
synchronized proprioceptive state feedback and receives independently
verified robot-specific commands at every decision cycle.

The architecture maintains a strict separation between semantic
reasoning and motor command authority. Qwen2.5-VL provides structured
scene evidence and Qwen3 produces symbolic task-level proposals; both
language models function as advisory components and cannot authorize
physical motion. Command authority rests exclusively with the Python
zonotope reachability module, which propagates robot-specific
reachable tubes and evaluates obstacle avoidance, safe-corridor
containment, and inter-robot separation predicates before any command
is published through MQTT. Commands that fail any predicate, or that
arrive as stale packets beyond the validity horizon, are withheld
and a conservative directive is issued instead.

Online validation under controlled indoor laboratory conditions demonstrated
that the pipeline approves motion in clear-path scenarios, redirects candidate
paths around dynamic human obstacles when a geometrically feasible corridor
remains, and withholds commands when any reachability predicate or freshness
condition is violated. Quantitatively, over $1{,}953$ consecutive decision
cycles ($65.1$\,s at $30$\,Hz), $973$ object detections and $17{,}577$ verified
candidate arcs, the gate approved motion in $39.8\%$ of cycles and withheld it
in $60.2\%$. Per-scenario approval spanned $56.4\%$ in the clear-path office
approach to $0.0\%$ at the non-traversable stairwell, where corridor
containment alone rejected all $3{,}249$ candidate arcs while the obstacle
predicate was never violated. Internal \textsc{safe} occupancy fell from
$92.2\%$ to $18.8\%$ as obstacle proximity increased, while decision output
remained stable at a $0.476$\,Hz switching rate against the $30$\,Hz cycle
rate. Decision-conditioned statistics confirm geometric rather than semantic
gating: \textsc{proceed} was issued at mean risk $0.155$ and $2.43$\,m
clearance, \textsc{avoid+proceed} at mean risk $0.737$ and $0.92$\,m clearance
against a pedestrian detected as close as $0.51$\,m. The compound inflation
model was validated quantitatively: semantic confidence explains $9.1\%$ of
cycle-risk variance ($r = 0.301$, $n = 794$) against $36.8\%$ for obstacle
geometry ($r = -0.607$, $n = 855$), a factor of four that rules out
confidence-only safety margins. Against thirteen representative systems
compared on six capability axes in Table~\ref{tab:sota_comparison}, the
proposed pipeline is the only one providing constraint transfer to a platform
without exteroceptive sensing and an inter-robot separation predicate on paired
reachable tubes, with prior-art coverage at $0/12$ on both. A representative per-cycle latency budget of approximately
$1200\,\mathrm{ms}$ was derived for the laboratory configuration,
indicating that the implementation is appropriate for supervisory
low-speed indoor navigation. The pipeline demonstrates, under the
tested conditions, that strict separation between semantic reasoning
and command authority, combined with robot-specific zonotope
reachability gating, supports safer coordination of heterogeneous
robots with asymmetric sensing. It does not constitute a universally
safe controller, and generalization beyond the tested conditions
requires additional systematic validation.

Future work will target larger and more complex indoor and outdoor
environments, teams with more than two robots, longer trial durations
with more diverse dynamic obstacles, improved camera calibration and
stronger uncertainty models for more accurate obstacle inflation,
decentralized server architectures for fault tolerance, hardware-in-the-loop
and long-duration operational experiments, instrumentally measured
latency benchmarking over extended runs, scalability analysis for
three or more robots, and evaluations oriented toward safety
certification criteria.

\nocite{*}
\bibliographystyle{IEEEtran}
\bibliography{ref}

% ============================================================
%  Author Biographies
% ============================================================
\begin{IEEEbiography}[{\includegraphics[width=1in,height=1.25in,clip,
keepaspectratio]{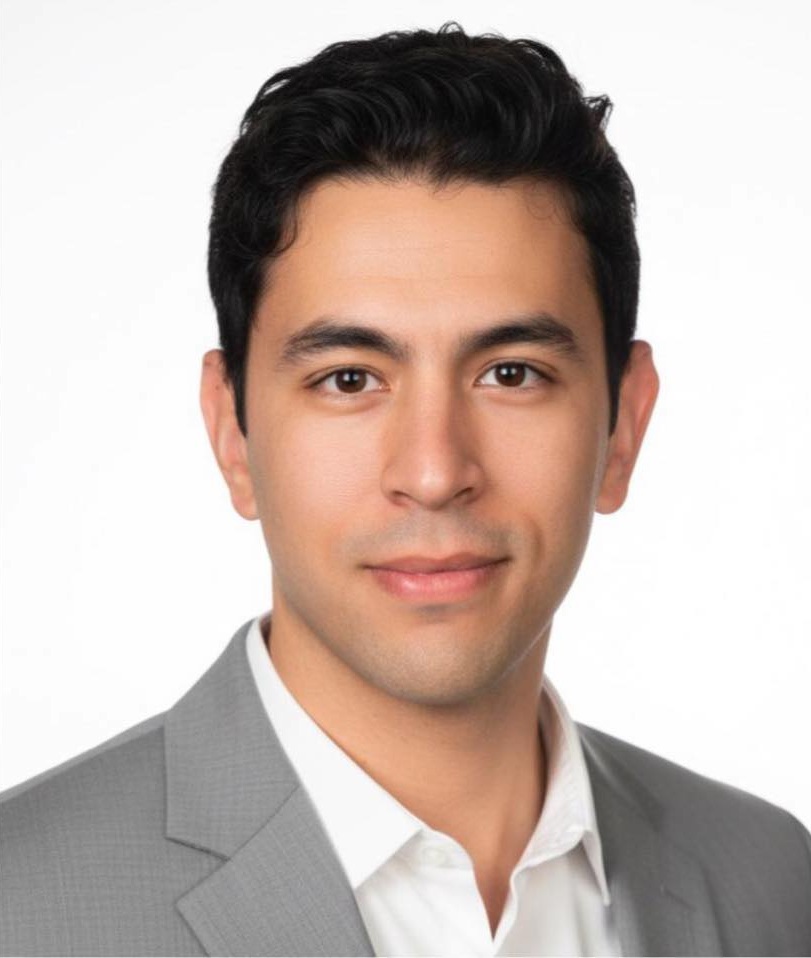}}]{Mohamed Dwedar}
received  two M.Sc. degrees, one in Mechanical Engineering from Selçuk University, Turkey, and one in Mechatronics and Artificial Intelligence from the University of Siegen, Germany. He is currently pursuing the Ph.D. degree in Cyber-Physical Systems with the Professorship of Cyber-Physical Systems (Heilbronn), Department of Computer Engineering, TUM School of Computation, Information and Technology, Technische Universität München (TUM), Germany. His doctoral research spans machine learning, multi-agent systems, and intelligent robotics. Prior to his doctoral studies, he held research positions at Bonn-Rhein-Sieg University of Applied Sciences and Cologne University of Applied Sciences, where he developed machine learning models for mechatronics systems optimization. He also contributed to the 5G-Bridges project on anomaly detection in machine-to-machine (M2M) communications and the KILaserCleaner project on AI-driven decision-making for laser cleaning applications. His research interests include agentic AI, vision-language models, real-time IoT communication, and autonomous multi-robot coordination encompassing unmanned aerial vehicles and ground robotic platforms. He has authored multiple publications in IEEE and Springer venues and has contributed book chapters in Artificial Intelligence and Robotics. He was a Global Winner of the NASA Space Apps AI Competition in 2024.
\end{IEEEbiography}

\begin{IEEEbiography}[{\includegraphics[width=1in,height=1.25in,clip,
keepaspectratio]{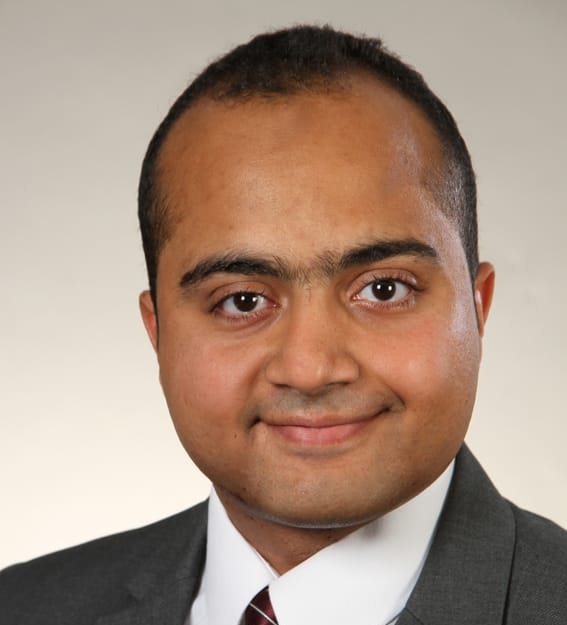}}]{Ahmad Hafez}
received two M.Sc. degrees, one in Energy Engineering from Technische Universität Berlin, and one in Data Analytics from Stiftung Universität Hildesheim, Germany. He is currently a Graduate Research Fellow pursuing the Ph.D. degree with the Professorship of Cyber-Physical Systems (Heilbronn), Department of Computer Engineering, TUM School of Computation, Information and Technology, Technische Universität München (TUM), Germany, working under the supervision of Prof. Amr Alanwar. With over 15 years of experience spanning energy systems, industrial AI, and formal methods, he previously held technical roles at Bosch, Siemens, and ABB. His research focuses on the intersection of formal verification and intelligent autonomous systems, with particular emphasis on data-driven reachability analysis, safe LLM-controlled robotics, and graph-based anomaly detection in cyber-physical systems. His work on constrained polynomial logical zonotopes, published in IEEE Control Systems Letters in collaboration with Prof. Karl H. Johansson (KTH Royal Institute of Technology), advances computationally efficient formal guarantees for logical systems.\end{IEEEbiography}

\begin{IEEEbiography}[{\includegraphics[width=1in,height=1.25in,clip,
keepaspectratio]{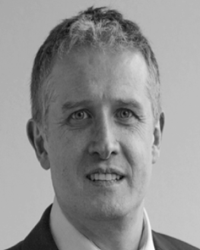}}]{Alexander Jesser}
received the Diploma degree in Computer Engineering from the University of Paderborn, Germany, and the Ph.D. degree in Computer Engineering from Johann Wolfgang Goethe University of Frankfurt am Main, Germany. Prior to his academic career, he held leading positions in the medical technology and automotive industries. Since 2013, he has been a Full Professor in Embedded Systems and Communications Engineering with Heilbronn University of Applied Sciences, Germany, where he founded and has headed the Institute of Intelligent Cyber-Physical Systems (ICPS) since 2021. His research interests include cyber-physical systems, signal and image processing, voice processing, and their applications in industrial and medical technology. He has served on several international scientific committees, including RICOTED, and has held Visiting Professor positions at Paraguayan-German University (UPA), Asuncion, Shenzhen University of Technology (SZTU), and Dulaty University, Kazakhstan. In 2024, he was awarded an honorary Professorship from Paraguayan-German University (UPA) in recognition of his outstanding international contributions.
\end{IEEEbiography}

\begin{IEEEbiography}[{\includegraphics[width=1in,height=1.25in,clip,keepaspectratio]{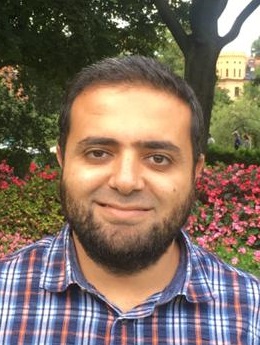}}]{Amr Alanwar} received the M.Sc. degree in Computer Engineering from Ain Shams University, Cairo, Egypt, in 2013, and the Ph.D. degree in Computer Science from the Technische Universität München (TUM), Germany, in 2020. He was a Postdoctoral Researcher at KTH Royal Institute of Technology, Sweden, and a Research Assistant at the University of California, Los Angeles, CA, USA. He is currently an Assistant Professor with the Department of Computer Engineering, TUM School of Computation, Information and Technology, Technische Universität München, Germany. He received the Emmy Noether Fellowship from the German Research Foundation (DFG) in 2025, the Best Paper Award in the Systems and Applications Track at HSCC/ICCPS during CPS Week 2026, and the Best Demonstration Paper Award at the IPSN during CPS Week 2017. He was a two-time finalist in the Qualcomm Innovation Fellowship.\end{IEEEbiography}

\EOD
\end{document}